\documentclass[12pt]{article}
\usepackage[T1]{fontenc}
\usepackage{times}
\usepackage[left=20mm,right=20mm,top=20mm,bottom=15mm,headsep=0pt]{geometry}

\usepackage{amsmath,amssymb,amsthm}

\usepackage{algorithm}
\usepackage{algcompatible}
\usepackage{graphicx}
\usepackage{caption}
\usepackage{subfigure}
\usepackage{multirow}
\usepackage[utf8]{inputenc}
\usepackage{comment}
\usepackage{colortbl}
\usepackage{color,soul}
\usepackage{ulem}
\usepackage[dvipsnames]{xcolor}
\usepackage{soul}
\usepackage{cite}
\usepackage{stfloats}

\newtheorem{theorem}{Theorem}
\newtheorem{lemma}[theorem]{Lemma}
\newtheorem{corollary}{Corollary}

\theoremstyle{definition}
\newtheorem{definition}{Definition}

\theoremstyle{definition}

\theoremstyle{definition}
\newtheorem{remark}{Remark}

\newcommand{\vctilde}[1]{\tilde{\mathbf{#1}}}
\newcommand{\set}[1]{\mathbf{#1}}

\begin{document}

	\begin{center}
			{\huge\textrm{Dynamic Network-Assisted D2D-Aided \\Coded Distributed Learning}}\\
			\vspace{0.75cm}
			\large{Nikita Zeulin$^{1}$, Olga Galinina$^{1}$, Nageen Himayat$^{2}$, Sergey Andreev$^{1}$,\\ and Robert W. Heath Jr.$^{3}$}\\
			\vspace{0.25cm}
			\normalsize{$^{1}$Tampere University, Tampere, Finland,  $^{2}$Intel Corporation, Santa Clara, CA 95054-1549, USA, \\$^{3}$North Carolina State University, Raleigh, NC, USA}\\
			\vspace{0.25cm}
			\normalsize{\{nikita.zeulin, olga.galinina, sergey.andreev\}@tuni.fi, \\nageen.himayat@intel.com, rwheathjr@ncsu.edu}
	\end{center}
	\vspace{0.3cm}
	\begin{abstract}
		Today, various machine learning (ML) applications offer continuous data processing and real-time data analytics at the edge of a wireless network. Distributed real-time ML solutions are highly sensitive to the so-called straggler effect caused by resource heterogeneity and 
		alleviated by various computation offloading mechanisms that seriously challenge the communication efficiency, especially in large-scale scenarios. To decrease the communication overhead, we rely on device-to-device (D2D) connectivity that improves spectrum utilization and allows efficient data exchange between devices in proximity.
		In particular, we design a novel D2D-aided coded federated learning method (D2D-CFL) for efficient load balancing across devices.
		The proposed solution captures system dynamics, including \textit{data} (time-dependent learning model, varied intensity of data arrivals), \textit{device} (diverse computational resources and volume of training data), and \textit{deployment} (varied locations and D2D graph connectivity). 
		To minimize the number of communication rounds, we derive an optimal compression rate for achieving minimum processing time and establish its connection with the convergence time. The resulting optimization problem provides suboptimal compression parameters, which improve the total training time. Our proposed method is beneficial for real-time collaborative applications, where the users continuously generate training data resulting in the model drift.
	\end{abstract}
	
\section{Introduction}

Emerging applications of collaborative machine learning~(ML) for vehicular~\cite{8648365,7895118}, factory~\cite{sathyan2020comparison}, and aerial automation~\cite{choi2019unmanned} leverage wireless infrastructure to facilitate distributed training and data exchange. 
Examples of such applications include simultaneous localization and mapping~\cite{kegeleirs2021swarm} or cooperative sensing~\cite{8950168}, where multiple users jointly train a common model related to their operating environment. In such scenarios, wireless connectivity plays a fundamental role in supporting real-time collaborative training and inference.

To avoid excessive data transfers in the network, collaborative applications may rely on the principles of federated learning~(FL)~\cite{konecny2016federated}, which, however, is challenged by resource heterogeneity. Due to diverse communication and computation capabilities, some of the participants may impede the collaborative performance causing the so-called \textit{straggler} effect. Existing work to reduce the impact of stragglers includes mechanisms such as coded computing \cite{dhakal2019codedcomputing,prakash2020coded}, over-the-air computing \cite{8870236,8952884}, or specific resource allocation \cite{9210812,chen2020convergence,yang2020energy} and scheduling \cite{9170917}.

Another approach to improve spectrum utilization and communication efficiency, in general, is in device-to-device (D2D) offloading \cite{zhang2021d2d} that shifts the data exchange from the vertical user-server to a horizontal D2D plane, thus releasing the load on the cellular network~\cite{loghin2020disruptions}. While D2D connectivity facilitates efficient data exchange between the proximate devices, especially in highly heterogeneous systems with dense connectivity~\cite{9060868}, straightforward methods of information offloading
may violate user data privacy~\cite{7809147}.

To decrease the communication overhead while maintaining a certain level of data privacy, we propose a novel D2D-aided coded distributed learning method, where the network assists collaborative training and inference via a joint optimization of learning and offloading.Using an example of the least squares regression estimator, we provide a detailed assessment of the proposed method under dynamic conditions that include the model drift.

\subsection{Related Work} 

The straggler mitigation solutions in FL can be divided into communications and coding theory methods. An illustrative example of the former is \cite{9210812}, which proposes a joint resource and power allocation scheme that minimizes the overall training time subject to specific FL constraints, such as maximum delay, available bandwidth, computational capacity, energy consumption, and others. Alternatively, the work in \cite{9170917} excludes the stragglers following the policy designed to balance the iteration time and gradient divergence. Aggregation of partial results can become a bottleneck when the number of users exceeds the pool of channel resources. To overcome this constraint, \cite{8870236} introduces the broadband analog aggregation technique, which exploits natural additivity of the wireless channel and enables simultaneous transmission of the user outputs along with their summation directly in the air. 

The research in coded distributed ML approaches builds on the idea to distribute one common matrix-to-vector multiplication operation by finding an optimal computational load for each user, compressing the original data to its optimal size, and forwarding it to the users for further processing. This idea is adapted to FL in \cite{dhakal2019codedfederated} by solving a linear regression problem with distributed gradient descent. The method proposes offloading excessive computational load to a computationally strong server by finding an optimal load allocation. In reality, however, central servers that simultaneously handle multiple applications cannot always provide rich computational resources for each particular ML training task. 

The coding FL solution is further extended in \cite{prakash2020coded} by a privacy budget analysis and the use of random Fourier feature mapping \cite{rahimi2007random}, which allows predicting non-linear functions while preserving linearity of operations required by coding. The results in \cite{dhakal2019codedfederated} and \cite{prakash2020coded} demonstrate that coding significantly reduces the training times while achieving identical convergence in the number of iterations. Importantly, higher privacy budgets may induce slower convergence, which is reflected in a privacy--utility tradeoff as discussed, e.g., in~\cite{showkatbakhsh2018privacy,bartan2020distributed}. In our work, we address this fact by considering the choice of a compression rate minimizing the convergence time of the distributed online regression, which has not been covered by the existing coded distributed FL methods.

Data compression \cite{bartan2020distributed} applied to coded offloading is one of the ways to achieve \textit{differential privacy}, \cite{dwork2008differential}. The idea here is to add noise to the original data to minimize the risks of revealing a single data point if the rest of the data is compromised. Adding Gaussian or Laplace noise to the data to achieve differential privacy was extensively covered by the current background, and we direct the reader to 
\cite[Sec. II]{hassan2019differential} and \cite[Sec. 4]{kairouz2019advances}.
Though this approach became extremely popular for privacy preservation in FL, adding noise directly cannot accelerate data processing. In contrast, the encoding procedure we apply in our method introduces certain privacy guarantees while reducing computational and communications costs. Another approach for privacy-preservation in FL is applying cryptographic methods based on multi-party computing. These provide stronger privacy guarantees but usually require time-consuming decoding of the encrypted data before the actual processing. In our proposed solution, we resort to the privacy-aware encoding procedure that exploits properties of the underlying ML algorithm to allow computing over the encoded data directly and, therefore, to decrease the total training time.

While offloading to a computationally strong server enhances the computing performance and mitigates the straggler effect, the communication efficiency becomes the key bottleneck, which is further challenged by the growing number of participants and the size of the ML model. The main approaches to minimize the communication overhead include decreasing the size of the update at each iteration (e.g., using data compression or sketching~\cite{konecny2016federatedNIPS}) or reducing the total number of iterations~\cite{luping2019cmfl}. Alternatively, D2D connectivity can enhance spectrum utilization and network communication efficiency by decreasing the load of user-server data exchange~\cite{zhang2021d2d}. D2D communications are addressed in the survey \cite{loghin2020disruptions} and multiple related works as a potential mechanism for leveraging user proximity to improve FL solutions. However, the degree of information leakage over a D2D link remains largely unexplored. The recent work in \cite{9252973} proposes using direct communications for collaborative caching, however, without addressing the privacy aspects of D2D transmission explicitly. We argue that privacy is a key factor in collaborative D2D-aided computing~\cite{7809147} and aim at bridging the indicated gap in performance evaluation and optimization of the distributed learning process mindful of the user privacy.

\subsection{Main Contributions}

We propose a novel D2D-aided load balancing solution (termed D2D-CFL) for privacy-aware and communication-efficient distributed learning in highly dynamic environments. Our framework is capable of addressing the time and space heterogeneity of (i) the \textit{data} (time-dependent learning model, model drift, varied intensity of data arrivals), (ii) the \textit{device} (diverse computational resources and volume of training data), and (iii) the \textit{deployment} (varied locations and D2D graph connectivity). In more detail, the core contributions of this work can be summarized as follows: 

\begin{itemize}
	\item \textit{D2D-aided load balancing}. We introduce D2D-aided privacy-aware load balancing for real-time FL applications to decrease the network communication load and mitigate the impact of stragglers. 
	Specifically, our proposed solution is proven to reduce collaborative training times up to 50\% compared to the baseline FL scenario. The proposed method especially benefits from high computational heterogeneity and dense connectivity of the users. If the computation resources are homogeneous, the performance is identical to that of the baseline.
	\item \textit{Coded distributed learning}. We apply data compression using random Gaussian matrices to preserve privacy and reduce the volume of data offloaded over D2D links. We analyze the privacy budget of the proposed method using $\varepsilon$-\textit{mutual information differential privacy}, which is known to be stronger compared to $(\varepsilon,\!\delta)$-\textit{differential privacy}\,\cite{cuff2016differential}.
	\item \textit{Approximate convergence rate}. We assess the convergence rate of the proposed method and provide approximations for when the data is clean and fully compressed. We establish the optimal compression rate achieving the minimum processing time by solving a joint convergence rate and load allocation optimization problem.
	\item \textit{Performance analysis}. We support our analytical results with extensive system simulations. We vary computational heterogeneity level, user mobility, and model dynamics to demonstrate that if the parameters are chosen according to the proposed strategy, our solution outperforms the baseline at all times.
\end{itemize}

The remainder of this text is organized as follows. In Section II, we describe the distributed online learning scenario and introduce the key assumptions of our system model. In Section III, we provide a detailed analysis of the limitations and privacy aspects of the the proposed communication-efficient compression mechanism. In Section IV, we derive an algorithm achieving the optimal load allocation and solve the joint convergence rate and load allocation optimization problem. In Section V, we support our analysis by offering numerical results for several illustrative scenarios. In Section VI, we conclude by sharing the key outcomes and highlighting the most beneficial applications of the proposed solution.

\setlength\tabcolsep{0.5cm}
\begin{table*}[bp]\footnotesize
	\centering
	\caption{Summary of system parameters}
	\begin{tabular}{p{0.9cm}p{8cm}|}
		\hline
		\textbf{Notation} &\textbf{Description}\\
		\hline
		$N$ &Number of users\\
		$m$ &Total number of points per iteration\\
		$d$ &Number of features\\
		$\sigma$ &Noise of observations\\
		$\alpha$ &Learning rate\\
		$\set{X}$ &Data matrix, $\set{X}\in \mathbb{R}^{n \times d}$\\
		$\set{y}$ &Vector of observations, $\set{y} \in \mathbb{R}^{n}$\\
		$g$ & Estimated function\\
		$\boldsymbol{\beta}^{(t)}$ &Estimate of regression model at iteration $t$, $\boldsymbol{\beta}^{(t)} \in \mathbb{R}^{d}$\\
		$\set{G}$ &Compression matrix, $\set{G}\in \mathbb{R}^{c \times n}$\\
		$\tilde{\set{X}}$ &Matrix of compressed data, $\tilde{\set{X}}\in \mathbb{R}^{c \times d}$\\
		$\tilde{\set{y}}$ &Vector of compressed observations, $\tilde{\set{y}} \in \mathbb{R}^{c}$\\
		$f$ &Cost function (empirical risk)\\
		$\nabla \! f,~\nabla\! \tilde{f}$ &Gradient of $f$ for clean and compressed points\\
		$\varepsilon$ &Privacy budget\\
		$\ell_i$ &Batch size of user $i$ per iteration\\
		$\gamma_i$ &Compression rate of user $i$\\
		$c_i$ &Number of compressed points\\
		$\omega_i$ &Fraction of compressed points of user $i$\\
		$\hat{\ell}_i$ &Number of points to compress\\
		$a_i$ &Computational rate of user $i$ (operations per second)\\
		$T^*$ &Total processing time for all users, deadline\\
		$T_i$ &Processing time of user $i$\\
		$T_{i,\mathrm{gd}}$ &Gradient computation time\\
		$T_{i,\mathrm{cpr}}$ &Compression time\\
		$T_{i,\mathrm{cm}}$ &D2D transmission time\\
		$b_f$ &Size of floating number in bits\\
		$r_{\text{D}}$  &D2D transmission rate (bit/sec)\\
		$R_D$ &D2D proximity radius\\
		 $\mathcal{L}$ & Loss function\\
		\hline
	\end{tabular}
\end{table*}

\begin{figure}[!h]
	\centerline{\includegraphics[width = 0.3\textheight]{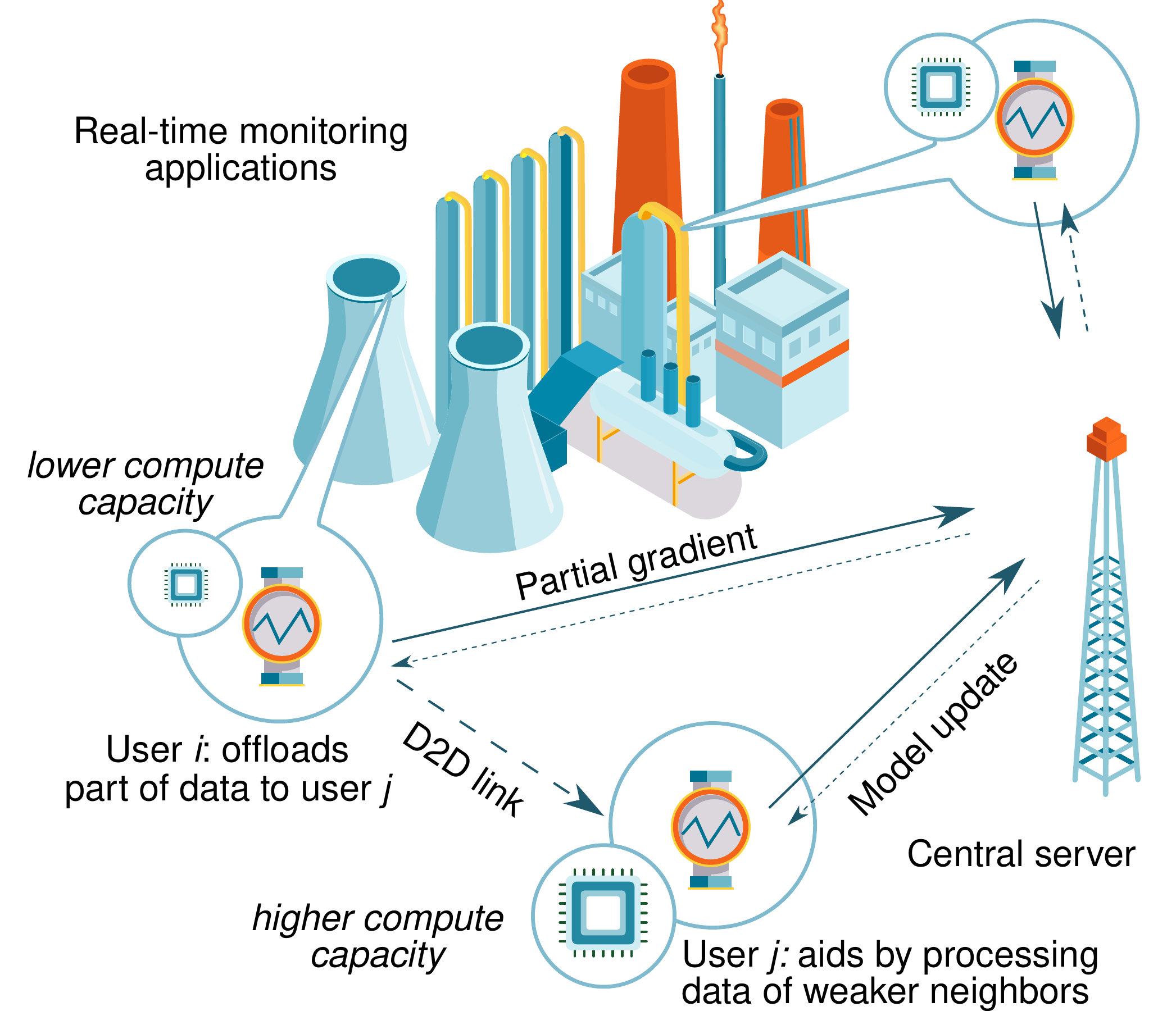}} \hspace{5px} 
	\caption{Illustration of D2D-aided distributed learning scenario, where two devices train collaborative model iteratively by exchanging updates with central server that aggregates partial gradients of all data points and orchestrates distributed learning. Device with lower computational capacity may offload excessive data to its stronger neighbor to accelerate training process. }
	\label{fig:main}
	\vspace{-0.4cm}
\end{figure}

\section{System Model}
In this section, we provide a detailed description of the online {learning} model, outline the key networking assumptions behind the scenario of interest, and introduce the target metrics employed for evaluating the performance of our framework. Hereinafter, vectors and matrices are given in bold as well as lowercase and uppercase, respectively, while non-bold lowercase variables are scalars.
\subsection{Scenario Description and Learning Model}

\textit{Regression model.} We consider a distributed online learning scenario, where a network of $N$ dynamic heterogeneous users, with the assistance of a central server (base station or access point), collaboratively solve a regression problem {(see example in Fig.~\ref{fig:main})}. Let $\set{X}\in\mathbb{R}^{m\times d}$ denote a combined dataset that consists of $m$ feature vectors $\set{x}_{j} \in \mathbb{R}^{d}$ and corresponding observations $y_{j}\in\mathbb{R}$; particularly, $ m= \sum _{i=1}^{N} \ell_i$, where $\ell_i$ is the number of data points at the device $i$. In general, solving a regression problem is equivalent to selecting a function $g(\set{x}_{i})$ that approximates the response $y_{j}$ best. For a set of functions $g(\set{x}_{j}, \boldsymbol{\beta})$, the regression problem can be formulated as finding the parameter vector $\boldsymbol{\beta}^* \in \mathbb{R}^d$ that minimizes the empirical risk constructed on the basis of $\set{X},\set{y}$:
\begin{equation}
	\boldsymbol{\beta}^*= \arg \min_{\boldsymbol{\beta}\in\mathbb{R}^d} 
	\left[\sum_{j=1}^{m} \mathcal{L}(\boldsymbol{\beta};\set{x}_{j},y_{j})\right]
	\equiv
	\arg\min_{\boldsymbol{\beta}\in\mathbb{R}^d}f(\boldsymbol{\beta}),
	\label{eq:erm}
\end{equation}
where $\mathcal{L}(\cdot)$ is the loss function determined by the regression model, and $f(\boldsymbol{\beta})$ is the empirical risk multiplied by $m$ (i.e., the cost function). 
We assume that $ \mathbf{x}_{{j}}$ are normalized, i.e., $\mathbb{E}[\mathbf{x}_{{i}}]=\mathbf{0}$ and $\text{var}(x_{{j,k}}) = 1$, $x_{{j,k}}$ is an element of $\mathbf{x}_j$.

In practical models, e.g., artificial neural networks (ANNs), the loss function $\mathcal{L}(\cdot)$ and the corresponding cost function $f(\cdot)$ are in general non-convex. Nevertheless, the stochastic gradient descent (SGD) \cite{bertsekas1989parallel} remains one of the most popular methods for solving optimization problem \eqref{eq:erm} for such models. The main focus of our D2D-CFL framework is SGD, and as an example, we consider the least squares regression model with empirical risk $f(\boldsymbol{\beta}) = \frac{1}{2}\left[ \left\| \set{X}\boldsymbol{\beta} - \set{y} \right\|_2^2 \right]$. The model is generalized to a non-linear formulation by using kernel feature mappings that approximate corresponding kernels. With the feature mapping, the original data is projected to a higher dimensional space where linear problem formulations can be efficiently applied.

\textit{Online stochastic gradient descent.} 
The estimated dependency $g(\cdot)$ may vary over time so that the training data become outdated and may no longer be useful in estimating the current value. Therefore, we focus on an online learning setup, where the data samples $\set{x}$ and the corresponding observations $y$ are being streamed continuously.
We divide the overall training time into iterations having duration $T$. During each iteration $t$, user $i$ processes the data $\left\{\set{X}_i^{(t)}, \set{y}_i^{(t)} \right\}$ generated by the end of the previous iteration while simultaneously collecting new training data for the next iteration. Our further analysis holds for every iteration, and, hence, we omit any iteration indexing except for the cases where we specifically address two consecutive iterations. Therefore, by the beginning of the tagged iteration, user $i$ has generated $\ell_i$ 
data points $\set{X}_i=\left(\set{x}_{i,1}^\intercal,\ldots,\set{x}_{i,\ell_i}^\intercal\right)^\intercal$, $\set{X}_i\in\mathbb{R}^{\ell_i\times d}$, and the corresponding observations $\set{y}_i=\left(y_{i,1},\ldots,y_{i,\ell_i}\right)^\intercal$, $\set{y}_i\in\mathbb{R}^{\ell_i}$, {while the total generated data and the corresponding observations are $\set{X}=\left(\set{X}_1^\intercal,\ldots,\set{X}_N^\intercal\right)^\intercal\in\mathbb{R}^{m\times d}$ and $\set{y}=\left(\set{y}_1^\intercal,\ldots,\set{y}_N^\intercal\right)^\intercal\in\mathbb{R}^{m}$, respectively.}

{In our framework, optimization problem \eqref{eq:erm} is solved with SGD, which iteratively updates {the parameter vector} $\boldsymbol{\beta}^{(t)}$} until a stopping condition is met. The online 
{SGD} updates the {parameters of the regression model} using new training data at each iteration. In this work, we focus on the {batched version of SGD}, which utilizes the data batches collected during the previous iteration. At the end of $t$-th iteration, the estimate $\boldsymbol{\beta}^{(t+1)}$ is updated according to the rule $\boldsymbol{\beta}^{(t+1)}=\boldsymbol{\beta}^{(t)}-\alpha\nabla \!f^{(t)}$, where $\nabla \! f^{(t)}= \set{X}^{(t)\intercal}\left(\set{X}^{(t)}\boldsymbol{\beta}^{(t)}-\set{y}^{(t)}\right)$ is the gradient of {the least squares linear regression} cost function $f(\boldsymbol{\beta})$, and $\alpha$ is the learning rate. The choice of $\alpha$ {for the least squares linear regression model} is addressed below in subsection IV.C.

Noteworthy, we do not consider collecting data samples over several consecutive iterations since the {estimated function $g(\cdot)$} is allowed to gradually evolve in time and, thus, data aggregation may significantly deteriorate  the precision of the estimated model. {We also assume that the employed variables correspond to the current iteration $t$ unless noted otherwise, and therefore omit the iteration indexing.}

\subsection{Centralized Assistance}

We focus on the case of centralized network assistance, where the server is responsible for arbitrating the entire learning process, collecting the results of local calculations, and disseminating the updates. Particularly, at the beginning of each iteration $t$, the server broadcasts the global model $\boldsymbol{\beta}$ to the users. Then, $i$-th user computes a partial gradient
\begin{equation} \label{eq:s2:gradient_expression}
	\nabla \! f_i=\set{X}_i^{\intercal}\left(\set{X}_i\boldsymbol{\beta}-\set{y}_i\right)
\end{equation}
\noindent
and transmits it to the server. Here, index $(t)$ is omitted for clarity; however, all variables are time-dependent. The server aggregates partial gradients of the users, obtains a full gradient $\nabla \! f=\sum_{i=1}^N \nabla \! f_i$ of this iteration, and updates the global model. In our online learning scenario, we assume that the training process is continuous, and, thus, we do not introduce any stopping conditions.

At the beginning of an iteration, the server is fully aware of the states of the users, including their locations, numbers of collected data points, and computational capacities. The server may leverage this information to devise mechanisms of optimal load allocation that minimize the iteration time. Particularly, the server advises the users on the number of points to be processed locally or offloaded to the neighboring devices.

\subsection{Computational Resources}

In our setup, the basic computational operation is multiply-accumulate (MAC), {which computes the} inner product of two $d$-dimensional vectors{. One MAC is equivalent to $2d+1$ floating point operations and is used for compact and straightforward analytical expressions.} The computational power of $i$-th user $a_i$ is defined as the number of MAC operations per second, or the MAC rate (MACR). The values $a_i$ are constant during each iteration but may vary between them. 

\textit{Computational heterogeneity:} We assume that computational resources {vary across the users} and follow a certain distribution {that} defines {the heterogeneity level} in the system and, thus, affects the optimal load allocation and {iteration time.}

We consider a so-called \textit{multi-tenancy scenario}, where multiple services of different owners co-exist in a shared environment and may receive only limited support from the network. Under this scenario, the third-party intelligent edge applications are low-priority to the network server, which cannot allocate resources for demanding computational tasks. 

\textit{Server computational capacity:} {We assume that} the network assistance is limited to performing {only} basic operations, such as aggregation of partial gradients or updating and broadcasting the global model. {Hence, the computational capacity of the server cannot be allocated for gradient computations and is assumed to be equal to zero.}

\subsection{Deployment and Direct Connectivity}
{The regression model is trained distributedly by the users, which move within a certain area of interest}. 

\textit{D2D assistance:} The users are allowed to establish D2D connections with their neighbors within the proximity radius $R_D$ fixed across the network. The parameter $R_D$ is selected by the server based on the path loss information and the chances that D2D link availability would persist throughout the iteration time. We define a set of users $\mathcal{N}(i)$ located within the distance of $R_D$ from $i$-th user. The transmission rate $r_\text{D}$ over a direct link is constant and kept equal for all users. 

\textit{{Mobility during iteration}:} {We assume that the iteration time is sufficiently short so that the  resources and the set of available neighbors $\mathcal{N}(i)$ for $i$-th user do not change during one iteration. Therefore, one may assume that the server can reliably plan the operation based on the knowledge available at the beginning of each iteration.} 

\textit{Roles of participants.} In our setup, all users simultaneously gather new training data from the environment, e.g., using their built-in sensors, and process the data collected during the previous iteration. The users that have weaker computational capacities or higher numbers of data points to process may take advantage of {computational heterogeneity. Particularly, they may leverage direct connectivity to offload their encoded data to more capable neighbors while maintaining certain privacy guarantees of data transmission.} As a result, all participants act not only as data generators and processors but may also serve as helpers by calculating the gradients of points arriving from less capable proximate sources. 

We assume that the server continuously tracks the coordinates of mobile users and is aware of the user state in terms of the data arrivals $\ell_i$ and capacities $a_i$. Based on this information, the server orchestrates the interactions across the network, assists the users in establishing direct links, and controls the distributed learning process.

\subsection{Performance Metrics: Privacy and Convergence}
{We evaluate the performance of the proposed offloading strategy using two key performance indicators, $\varepsilon$-mutual information differential privacy and convergence of the optimization process.}

\subsubsection{Differential privacy} 
{We estimate the privacy budget $\varepsilon$ of our proposed solution using mutual information differential privacy, which allows for information-theoretic analysis while being stronger than commonly used $(\varepsilon,\delta)$-differential privacy \cite{dwork2006our}.}

\begin{definition}[$\varepsilon$-MI-DP, \cite{cuff2016differential}]
	Let $\mathbf{D}$ be a set of data points {drawn from a probability distribution $p_D(\mathbf{D})$}, $\mathbf{d}_j \in \mathbf{D}$ be $j$-th data point of $\mathbf{D}$, $\mathbf{D}_{-j} = \mathbf{D} \!~\backslash~\! \mathbf{d}_j$ be the set obtained from $\mathbf{D}$ by excluding  $j$-th point, and $Y$ be an arbitrary measurable set. 
	The random mechanism $q(\cdot): \mathbb{R} \to Y$ satisfies \textit{$\varepsilon$-mutual information differential privacy} ($\varepsilon$-MI-DP) if 
	\begin{equation} \label{eq:s2:midp_def}
		\begin{array}{c}
			\sup_{j,p_{D}(\mathbf{D})}I(\mathbf{d}_j;q(\mathbf{D})|\mathbf{D}_{-j}) \le \varepsilon.
		\end{array}		
	\end{equation}
\end{definition}
\noindent
In other words, (\ref{eq:s2:midp_def}) {determines the privacy budget of the mechanism and represents the upper bound of the amount of information that} can be obtained about $j$-th data point from knowing the transformed data $q(\mathbf{D})$ and the remaining data points $\mathbf{D}_{-j}$. 

\subsubsection{Convergence time} 
We evaluate the learning convergence by measuring the convergence time, which is the minimum time required to reach the target performance. For simplicity, as the performance function at iteration $t$, we consider the normalized error of the regression model $\frac{||\boldsymbol \beta^{(t)} -\boldsymbol \beta^{*}||_2}{||\boldsymbol \beta^{*}||_2}$, {and, thus, the convergence time is defined by $\min_t \frac{||\boldsymbol \beta^{(t)} -\boldsymbol \beta^*||_2}{||\boldsymbol \beta^*||_2} < \delta$. We note that all} our results easily extend to the normalized mean squared error (NMSE) or the coefficient of determination (R-squared). For illustrative purposes, we measure the convergence time in both iterations and {time units} since the results are observed to be qualitatively different depending on the {duration of one iteration}. {Conversion from iterations to seconds can be readily done by multiplying the number of iterations by one iteration time.}

\section{D2D-Aided Coded Federated Learning}

In this section, we outline our D2D-aided coded {distributed learning} solution for {efficient} and privacy-{aware} online learning over a wireless network. We derive upper and lower bounds on the compression rate and establish the optimal compression rate that minimizes the iteration time. We also assess the achievable privacy budget of the D2D-CFL by considering two cases: compressed data transmission over a D2D link and partial gradient transmission.

\begin{figure*}[!h]
	\centerline{\includegraphics[width = 0.7\linewidth]{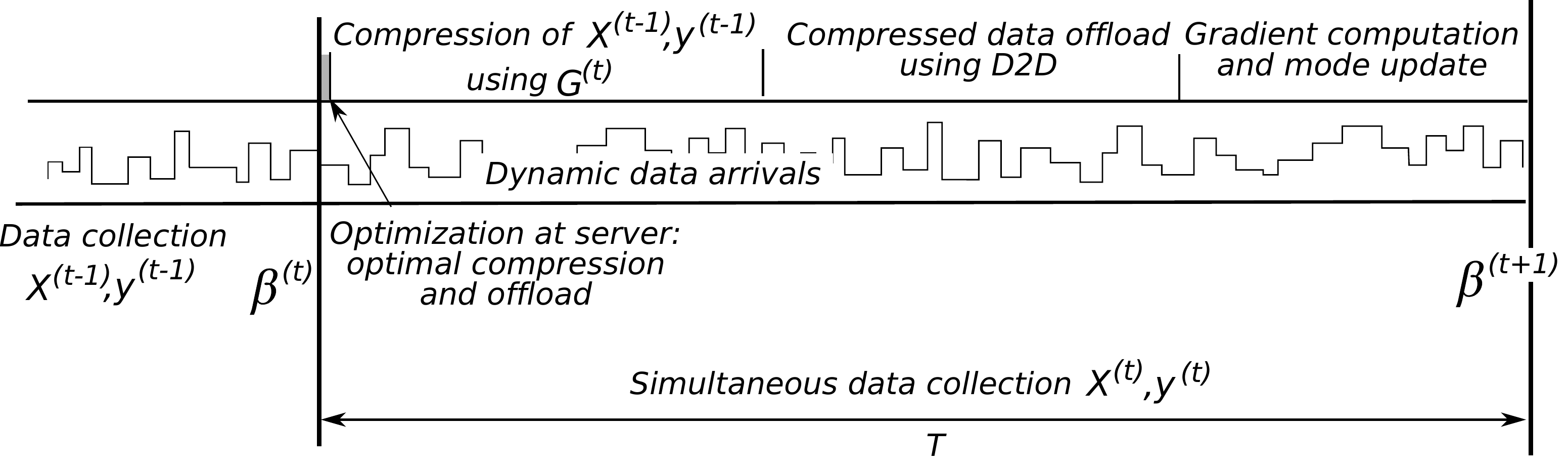}}
	\caption{{Time diagram for one iteration of D2D-CFL}.}
	\label{fig:s1:time_diagram}
\end{figure*} 

\subsection{Proposed D2D-CFL Solution}
To balance the computational workload in the system {and, thus, minimize the overall training time, we propose an approach based on the capability of users with lower computational resources to offload their excessive load to a stronger proximate device.} 
The straggling users compress part of their data with Gaussian coding matrices and then transmit the coded points to the devices advised by the network server that arbitrates the entire process. The stronger devices calculate the local gradients using their private uncoded data as well as the coded data shared by their weaker neighbors. Besides being differentially private \cite{showkatbakhsh2018privacy}, \cite{bartan2020distributed}, this approach may also reduce the iteration time by decreasing the number of points that participate in the calculation of local gradients.

We describe one iteration of the proposed D2D-CFL solution in more detail. We assume that the server may utilize information on the states of the users, including their positions, mobility models, numbers of data points, and computational capacities, {to devise an optimal computational load balancing scheme} that ensures fulfilling a preset deadline $T$. If users cannot meet the deadline $T$, they are considered \textit{weak} and are required to offload their data by compressing and transmitting them to the stronger neighbors. On the contrary, if users can meet the deadline $T$, they are considered \textit{strong} and are deemed potentially capable of receiving additional computational load to leverage their spare time before the deadline. Based on the state of the users, the server also calculates the optimal shares of data points to be offloaded; the optimal allocation is addressed in Section~IV. 

During each iteration, the (weak) users that are advised to share their coded data via D2D links perform the following three operations: data compression and coding, data exchange, computation of gradients (illustrated in Fig.~\ref{fig:s1:time_diagram}). After the users have transmitted or received the data points, the weak users compute the partial gradient of the remaining data points. The strong users compute the partial gradient of their local data points and the partial gradients of the received compressed data points and weigh the results with the appropriate learning rate $\alpha_i$. The latter is selected separately for each user depending on the number of uncoded and coded points. At the end of an iteration, the server aggregates the partial gradients, updates the model, and broadcasts the update to all the participants.

As a \textit{baseline} solution, we select a distributed learning approach, wherein the data is processed locally at the devices, and the iteration time is bounded by the processing time of the weakest user. We note that although the convergence time \textit{in iterations} may not be higher than that for D2D-CFL, we demonstrate that our solution is capable of providing a load allocation with notably lower convergence time \textit{in time units}.

\subsection{Fundamental Operations and Time Costs}

To keep the complexity of our system model reasonable, we assume that the time required for the server to aggregate the partial gradients and broadcast the global model is negligibly small, and, thus, we do not include it as part of the total processing time. To incorporate this service time, one may integrate it into the interval $T$.  {Next}, we list the corresponding time costs based on the MAC complexities $C(\cdot)$ of the gradient calculation, compression, and data transmission. 

\subsubsection{Data compression}

{Let $\gamma_i$ is the compression rate of $i$-th user, and $c_i=\gamma_i \ell_i$ be the number of compressed points of $i$-th user.}
We define the compression matrix of $i$-th user $\mathbf{G}_i$ as a $c_i \times \ell_i$ real-valued matrix, elements of which are drawn from the normal distribution with zero mean and variance of $1/\sqrt{c_i}$. Hence, the compressed data and the compressed observations are calculated as
\begin{equation}\label{eq:s2:compression_expression}
	\begin{array}{c}
		\tilde{\set{X}}_i=\set{G}_{i}\set{X}_i, \quad \vctilde{y}_i=\set{G}_i\set{y}_i.
	\end{array} 	
\end{equation}
\noindent
The gradient of the compressed data can be expressed as
\begin{equation}
	\label{eq:s2:compressed_gradient}
	{\nabla} \! \tilde f_i = \tilde{\set{X}}_i^{\intercal}\left(\tilde{\set{X}}_i\boldsymbol{\beta} - \tilde{\set{y}}_i\right) = \set{X}^\intercal_i\set{G}_i^\intercal\set{G}_i\left(\set{X}_i\boldsymbol{\beta} - \set{y}_i\right).
\end{equation}

The MAC complexities of data and observations compression are $C(\set{G}_{i}\set{X}_i)=c_i \ell_i=\gamma_i \ell_i^2$ and $C(\set{G}_{i}\set{y}_i)=c_i \ell_i/d = \gamma_i \ell_i^2 / d$, respectively.  We define the MAC complexity of a compression operation as the sum of $C(\set{G}_{i}\set{X}_i)$ and $C(\set{G}_{i}\set{y}_i)$, therefore, $C(\set{G}_{i}\set{X}_i,\set{G}_{i}\set{y}_i)=(1+1/d)\gamma_i \ell_i^2$. The time required to compress $\ell_i$ data points and observations to $\gamma_i \ell_i$ compressed points and observations is given by
\begin{equation}
	\label{eq:s2:Tcpr}
	\begin{array}{c}
		T_{i,\mathrm{cpr}}(\ell_i)=\left(1 + \frac{1}{d}\right) \gamma_i \ell_i^2 a_i^{-1}.
	\end{array}	
\end{equation}

\subsubsection{Gradient computation}

The MAC complexity of the operation in (\ref{eq:s2:gradient_expression}) is $C(\nabla \! f_i) = 2 \ell_i$. Hence, the time required to compute the gradient of $\ell_i$ data points is 
\begin{equation}
	\label{eq:s2:Tgd}
	\begin{array}{c}
		T_{i,\mathrm{gd}}(\ell_i)=2\ell_ia_i^{-1}.
	\end{array}	
\end{equation}

\subsubsection{Data transmission}

For the sake of consistency, we assume that $r_\text{D}$ is given in MAC/s. The transition to transmission rate $\hat{r}_\text{D}$ in bit/s can be done as $\hat{r}_\text{D}=db_fr_\text{D}$, where $b_f$ is a size of a floating point number in bits. Therefore, the time required to transmit $\ell_i$ data points and observations is 
\begin{equation}
	\label{eq:s2:Tcm}
	\begin{array}{c}
		T_{i,\mathrm{cm}}(\ell_i)=\left(1+\frac{1}{d}\right)\ell_ir_D^{-1}.
	\end{array}
\end{equation}

\begin{remark}
The application scope of the compression method \eqref{eq:s2:compression_expression}, also known as sketching, is not limited to the least squares linear regression. Recently proposed Neural Tangent Kernel \cite{jacot2018neural} allows representing the training of wide ANNs as a kernel regression problem with a convex least squares loss. In turn, the method in \cite{brand2020training} accelerates the kernel regression by reducing the complexity of the second-order optimization via fast sketching of the Jacobian matrix. Our D2D-CFL framework can be potentially extended to the distributed training of ANNs while enjoying the straggler mitigation via privacy-enabling data compression, as discussed below. 
\end{remark}

\subsection{Limitations of Data Compression}

{Below, we sequentially consider each stage of our D2D-CFL method in detail}. We analyze two essential aspects of data compression: the achievable compression bounds and the differential privacy introduced by compression. 

\subsubsection{Bounds on compression rate}

By utilizing the compression property, we (i) decrease the transmission time by reducing the number of transmitted data points, and (ii) may potentially decrease the data processing time by reducing the total volume of points in the system. While the first statement is intuitive, let us comment on the second one. Consider the case where $i$-th user compresses and forwards $\hat{\ell}_i=\omega_i \ell_i$ points to reduce the gradient computation time, where $\omega_i$ is the share of points to be compressed. {In the baseline approach, where partial gradients are computed locally, the processing time is based only on the calculation of the gradient, i.e., $T_{i,\mathrm{gd}}(\ell_i)$.} Then, the compression and forwarding the compressed points should not exceed {the local processing time}, which is formally
\begin{equation}\label{eq:s3a:gd_cpr_1}
	T_{i,\mathrm{gd}}(\ell_i-\hat{\ell}_i)+T_{i,\mathrm{cpr}}(\hat{\ell}_i)+T_{i,\mathrm{cm}}(\gamma \hat{\ell}_i) \le T_{i,\mathrm{gd}}(\ell_i),
\end{equation}
or
\begin{equation}\label{eq:s3a:gd_cpr_2}
	T_{i,\mathrm{gd}}(\hat{\ell}_i) - T_{i,\mathrm{cpr}}(\hat{\ell}_i) - T_{i,\mathrm{cm}}(\gamma \hat{\ell}_i) \ge 0.
\end{equation}

The expression in (\ref{eq:s3a:gd_cpr_1}) implies that the sum of times required to compress $\hat{\ell}_i$ points $T_{i,\mathrm{cpr}}(\hat{\ell}_i)$, offload $\gamma_i \hat{\ell}_i$ compressed points $T_{i,\mathrm{cm}}(\gamma \hat{\ell}_i)$, and compute gradient of $\ell_i-\hat{\ell}_i$ remaining points $T_{i,\mathrm{gd}}(\ell_i-\hat{\ell}_i)$ should not exceed the time required to compute the gradient of $\ell_i$ points $T_{i,\mathrm{gd}}(\ell_i)$ without performing  any offloading. Substituting the expressions (\ref{eq:s2:Tcpr})-(\ref{eq:s2:Tcm}) for $T_{i,\mathrm{cpr}}$, $T_{i,\mathrm{gd}}$, and $T_{i,\mathrm{cm}}$ into (\ref{eq:s3a:gd_cpr_2}), we may obtain an \textit{upper bound} on the maximum compression rate given the number of transmitted points $\hat{\ell}_i$ as
\begin{equation}\label{eq:s3a:gamma_ub_1}
	0 < \gamma \le \left(0.5(1+1/d)(\hat{\ell}_i + a_i/r_D)\right)^{-1}.
\end{equation}

{Since the compression mechanism produces an integer number of points, the offloaded data should be compressed to at least one point. Hence, we obtain} a trivial \textit{lower bound}: 
\begin{equation}
	\label{eq:s3a:gamma_lb}
	\gamma \ge \frac{1}{\hat{\ell}_i}.
\end{equation}

{Based on the above discussion, we can formulate the following Lemma.}
\begin{lemma}[D2D-CFL optimal compression rate] The optimal compression rate for $i$-th user {minimizing one iteration time} of D2D-CFL is given by
	\label{lemma:optimal_gamma}
	\begin{equation}
		\label{eq:s3a:optimal_gamma}
		\hat{\gamma}_i = \frac{1}{\hat{\ell}_i}=\frac{1}{\omega_i \ell_i},
	\end{equation}
	\noindent
	where $\omega_i$ is the share of compressed points. Hence, the number of compressed points of any user is $c = 1$. 
\end{lemma}

\begin{proof}
	{Consider a scenario where the channel conditions are ideal and $r_D\to\infty$. Substituting the latter to the upper bound~\eqref{eq:s3a:gamma_ub_1}, we obtain $0 < \gamma \le \left(0.5(1+1/d)\hat{\ell}_i\right)^{-1}$. One can readily see that the multiplier $2(1+1/d)^{-1}$ is bounded between $1$ for $d=1$ and $2$ for $d\to\infty$. Since the number of compressed points is integer and its upper bound is achieved only in asymptotic case, any number of transmitted points $\hat{\ell}_i$ is always compressed into one with the compression rate \eqref{eq:s3a:optimal_gamma}.}
\end{proof}

After deriving the bounds for the compression rate, we continue by assessing the differential privacy budget of the proposed solution.

\subsubsection{Differential privacy}

In our D2D-CFL framework, the wireless channel is exploited in two cases: to share the data points over a D2D link and to transmit the partial gradient to the server. We tailor the results of \cite{showkatbakhsh2018privacy} to analyzing the privacy budget of a direct link, where compression {mechanism \eqref{eq:s2:compression_expression}} is used to achieve {a certain} privacy level. 

\begin{lemma}[D2D-CFL direct link privacy budget]
	The D2D link privacy budget {introduced by the data compression} is given by
	\begin{equation}
		\varepsilon_i \geq \frac{1}{2}\log_2 \left(1 + \frac{1}{\omega_i \psi(\mathbf{X}_i)}\right).
	\end{equation}
\end{lemma}
\begin{proof}
	The employed coding and compression transform in (\ref{eq:s2:compression_expression}) achieves $\varepsilon$-MI DP if
	\begin{equation}\label{eq:s3:fX}
		\psi(\set{X}_i) \ge \frac{\gamma_i \ell_i}{2^{2\varepsilon_i}-1},
	\end{equation}
	where $\psi(\set{X})=\min_{j\in[1\ldots d]}\left\{\sum_{i=1}^l x_{i,j}^2 - \max_i x_{i,j}^2\right\}$~\cite{showkatbakhsh2018privacy}. Rearranging (\ref{eq:s3:fX}), we obtain
	\begin{equation}\label{eq:s3:gamma_dp_1}
		\gamma_i \le \frac{\psi(\set{X}_i)(2^{2\varepsilon_i}-1)}{\ell_i} \text{ and }
		2^{2\varepsilon_i} \ge \frac{\gamma_i\ell_i}{\psi(\set{X}_i)} + 1.
	\end{equation}
	
	Calculating the logarithm of \eqref{eq:s3:gamma_dp_1}, we express the privacy budget $\varepsilon_i$ as 
	\begin{equation}
		\label{eq:s3a:dp_budget_bound}
		\varepsilon_i \ge \frac{1}{2}\log_2 \left(1 + \frac{\gamma_i \ell_i}{\psi(\mathbf{X}_i)}\right).
	\end{equation}
	Substituting the optimal compression rate (\ref{eq:s3a:optimal_gamma}) into (\ref{eq:s3a:dp_budget_bound}), one may establish 
	\begin{equation*}
		\varepsilon_i \geq \frac{1}{2}\log_2 \left(1 + \frac{1}{\omega_i \psi(\mathbf{X}_i)}\right).
	\end{equation*}
\end{proof}
\begin{corollary}
	{Let $\set{x}_{u,v}$ be $(u,v)$-th element of $\set{X}_i$ and $\set{X}^{-(u,v)}$ be the set $\set{X}_i^{-(u,v)}$ without the element $\set{x}_{u,v}$.} The privacy budget of compressed gradient transmission is 
	\begin{equation*}
		\max\limits_{p(\set{X}_i)} I(\set{x}_{u,v};\nabla \tilde{f}_i\ |\ \set{X}_i^{-(u,v)}) \le \varepsilon_i.
	\end{equation*}
\end{corollary}
\begin{proof}
	This corollary follows directly from the data processing inequality $I(\set{x}_{u,v}; \set{G}_i\set{X}_i~|~\set{X}_i^{-(u,v)}) \ge I(\set{x}_{u,v}; \nabla \tilde{f}_i~|~\set{X}^{-(u,v)})$.
\end{proof}
\begin{remark}
	The compression mechanism alone does not necessarily yield {the privacy budget required by the users or the target application.} To achieve the required privacy level, one may additionally apply sequential composition \cite{hassan2019differential} of differentially-private mechanisms, such as Laplace or Gaussian.
\end{remark}

\begin{figure*}[!h]
	\centering
	\subfigure
	{
		\includegraphics[width=0.5\linewidth]{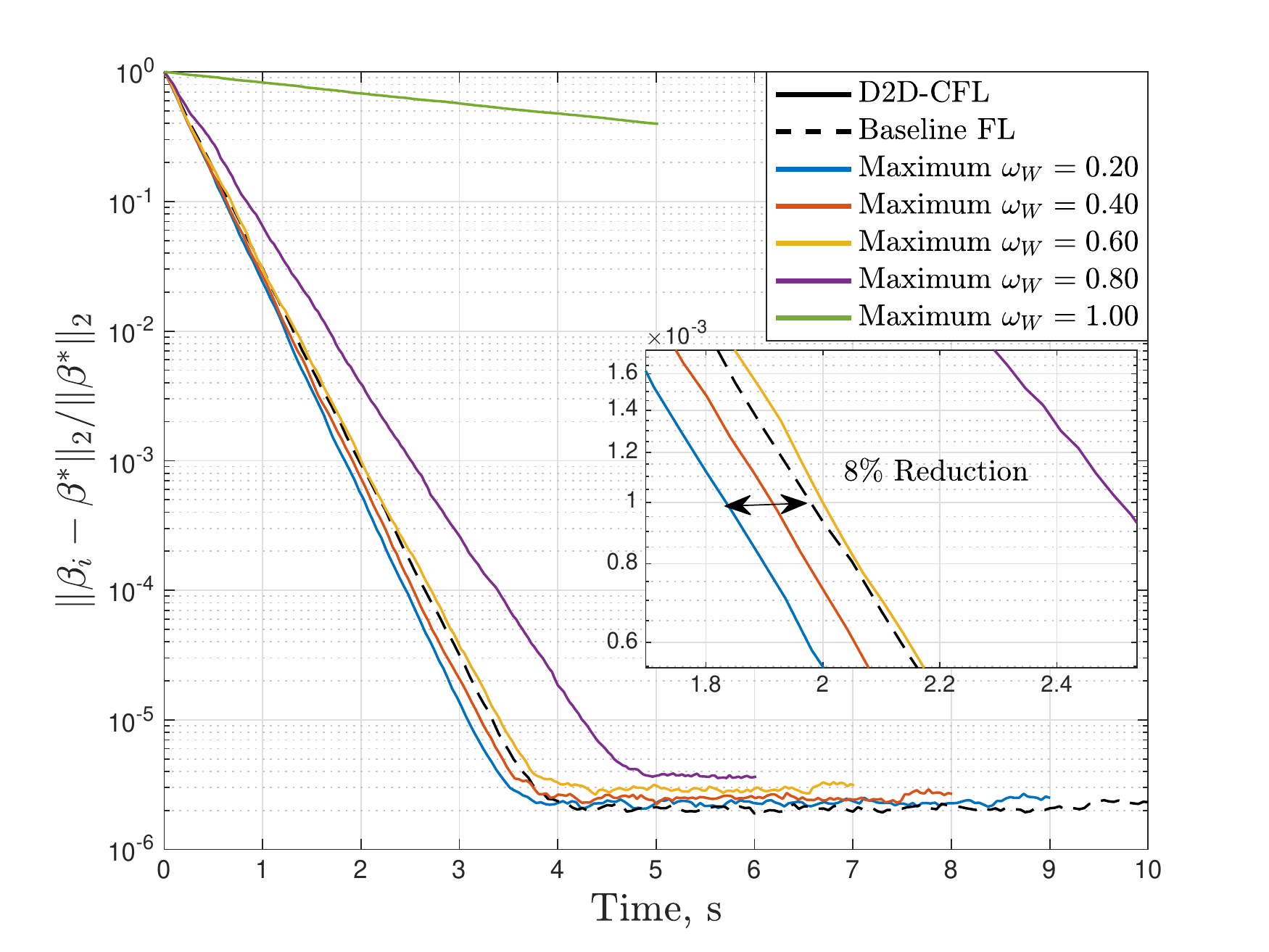}
	}~
	\subfigure
	{
		\includegraphics[width=0.5\linewidth]{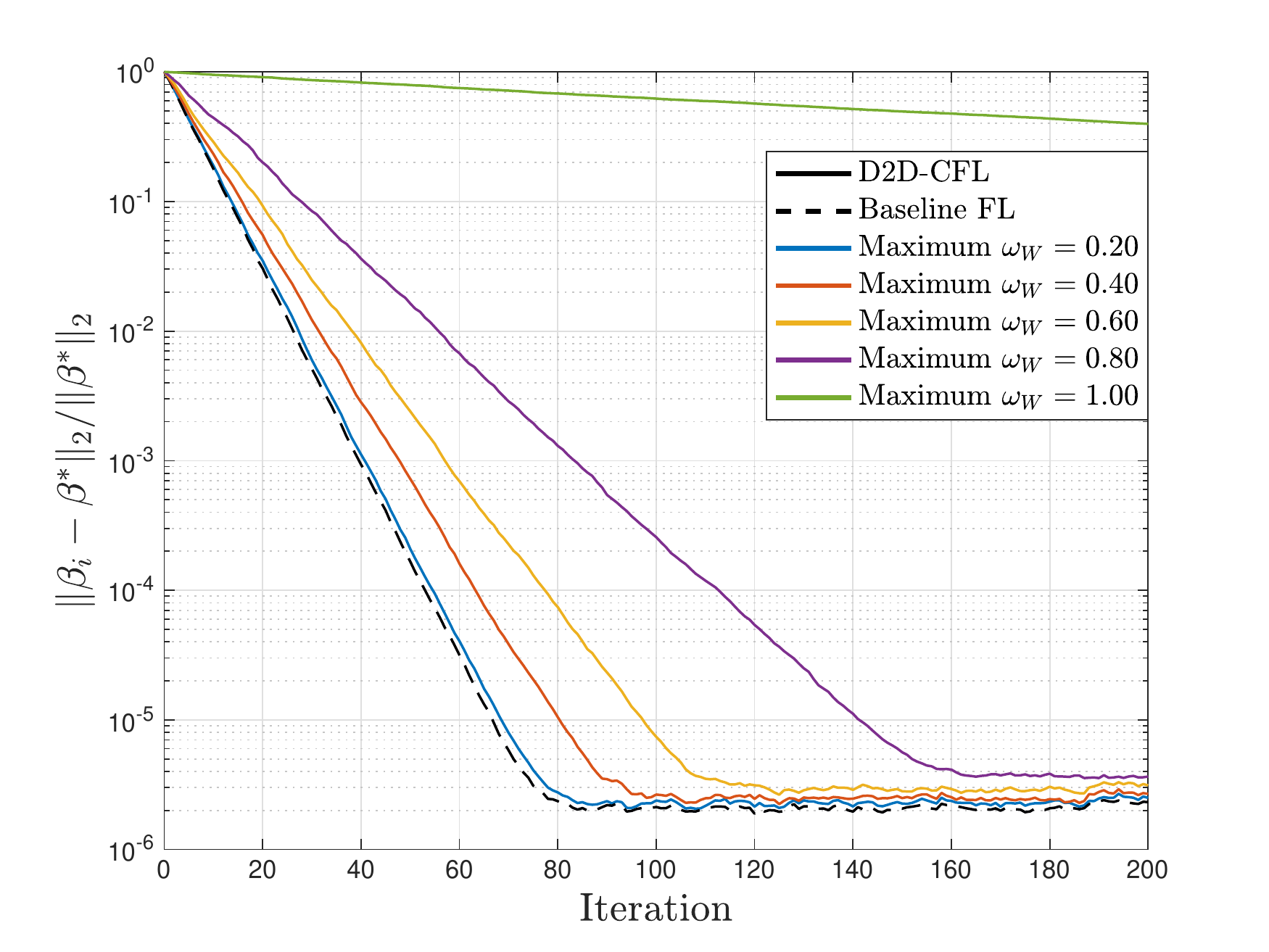}
	}
	\caption{Comparison of baseline FL and D2D-CFL in time and iterations. Static model $\boldsymbol{\beta}^*$: normalized error $\frac{||\boldsymbol{\beta}^{(t)} - \boldsymbol{\beta}^*||_2}{|| \boldsymbol{\beta}^*||_2}$ for \textit{equal} compression rate $\gamma_i$. Improvement is up to $8\%$.}
	\vspace{-3mm}
	\label{fig:eq_coding_synth}
\end{figure*}

\begin{figure*}[!h]
	\centering
	\subfigure
	{
		\includegraphics[width=0.5\linewidth]{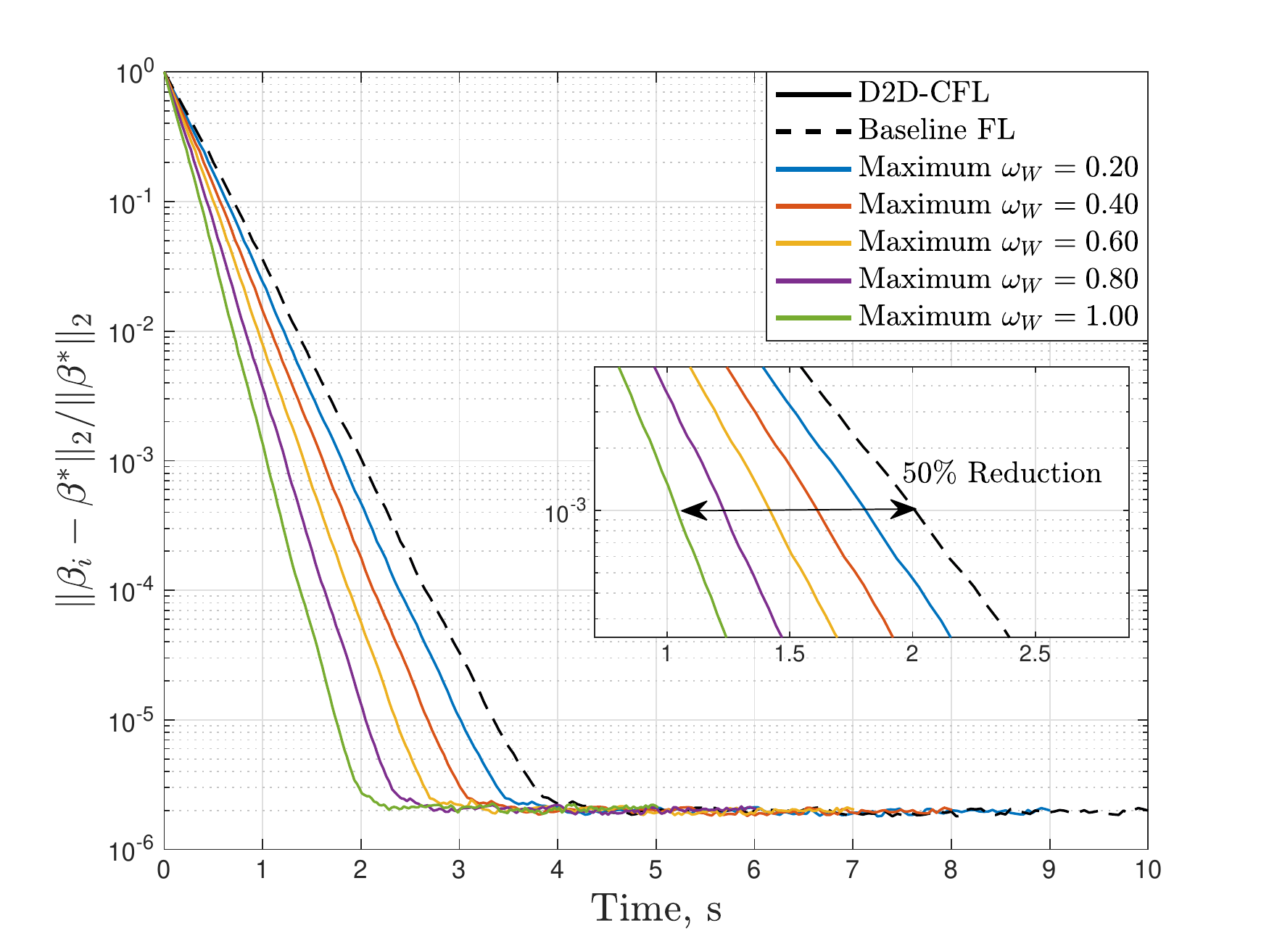}
	}~
	\subfigure
	{
		\includegraphics[width=0.5\linewidth]{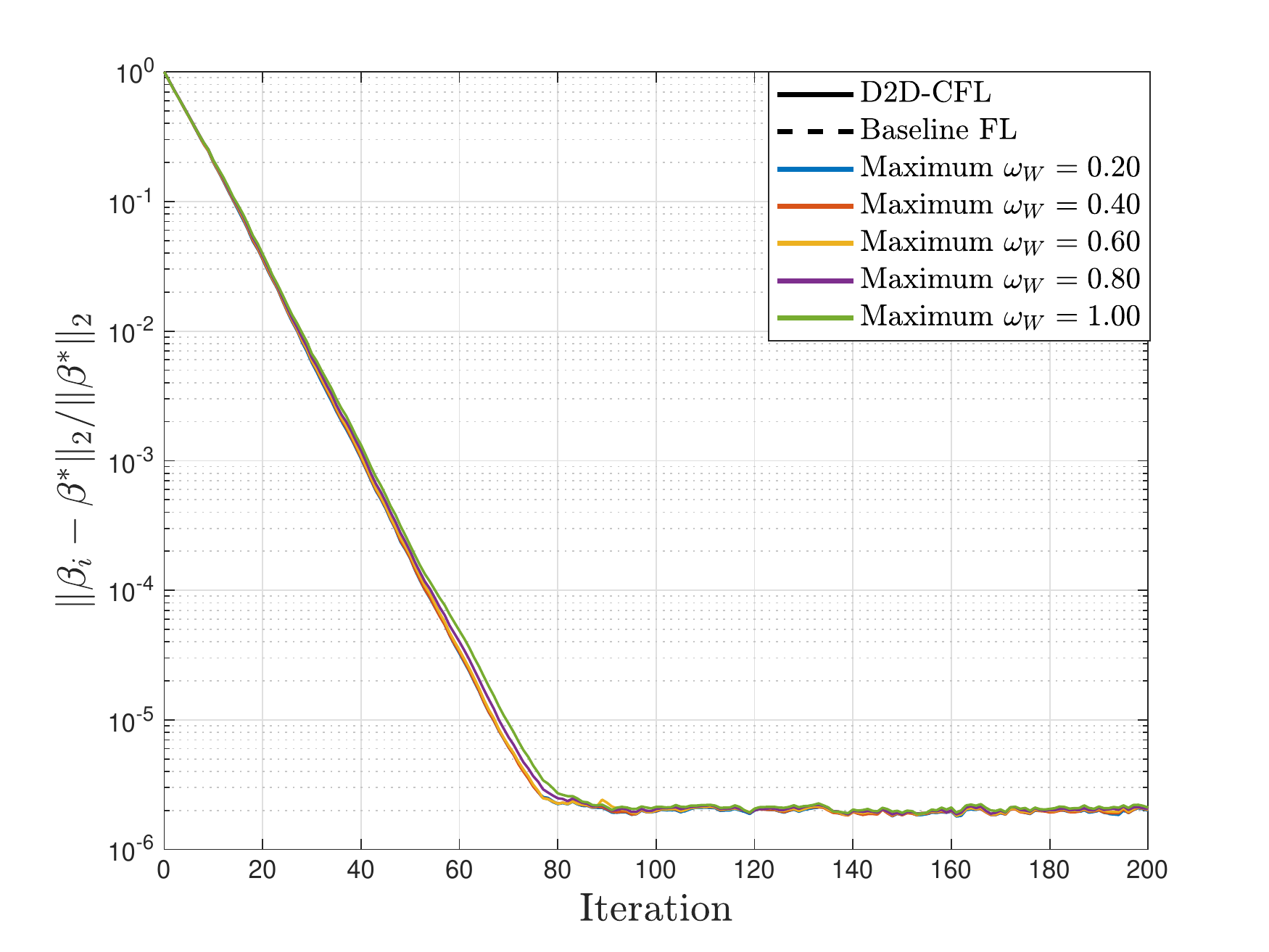}
	}
	\caption{Comparison of baseline FL and D2D-CFL in time and iterations. Static model $\boldsymbol{\beta}^*$: normalized error $\frac{||\boldsymbol{\beta}^{(t)} - \boldsymbol{\beta}^*||_2}{|| \boldsymbol{\beta}^*||_2}$ for \textit{optimal} compression rate $\gamma_i$. Improvement is up to $50\%$.}
	\vspace{-3mm}
	\label{fig:opt_coding_synth}
\end{figure*}

\section{Optimization of D2D-CFL Parameters}
\subsection{Optimal Load Allocation in D2D-CFL}
We propose a method for finding the optimal load allocation in terms of the total processing time. Let $T^*$ be the target deadline. Each user meets this deadline if 
\begin{equation}
	T^* \ge T_{i,\mathrm{gd}}(\ell_i - \omega_i \ell_i) + T_{i,\mathrm{cpr}}(\omega_i \ell_i) + \mathbb{I}(\omega_i > 0)T_{i,\mathrm{cm}}(1). \label{eq:Tstar_ineq}
\end{equation}
The inequality \eqref{eq:Tstar_ineq} can be {readily} expressed in terms of $T_{i,\mathrm{gd}}(\ell_i)$. Substituting $T_{i,\mathrm{gd}}(\ell_i)$ {by its definition \eqref{eq:s2:Tgd}} into {$T_{i,\mathrm{cpr}}(\ell_i)$ given by \eqref{eq:s2:Tcpr}}, we obtain $T_{i,\mathrm{cpr}}(\omega_i \ell_i)\!=\!\frac{1}{2}\left(1\!+\!\frac{1}{d}\right)\omega_i T_{i,\mathrm{gd}}(\ell_i)$, and $T_{i,\mathrm{gd}}(\ell_i \!-\! \omega_i \ell_i) \!=\! (1-\omega_i)T_{i,\mathrm{gd}}(\ell_i)$. {Then, the expression \eqref{eq:Tstar_ineq} can be represented as}
\begin{equation}
	\label{eq:s3a:time_gain}
	T^* \ge (1+ D \omega_i)T_{i,\mathrm{gd}}(\ell_i) + \mathbb{I}(\omega_i > 0)\left(2D+1\right)r_D^{-1},
\end{equation}
where $D = (0.5(1+1/d)-1) < 0$.
{Let us find share $\omega_i$ of points, which should be compressed and offloaded by $i$-th user to meet the preset deadline $T^*$}. Rearranging (\ref{eq:s3a:time_gain}) {as an equation}, we establish the optimal share of points as 
\begin{equation}
	\label{eq:w_i}
	\hat{\omega}_i = \left(\frac{T^* - T_{i,\mathrm{gd}}(\ell_i) - (2D+1)r_{D}^{-1}}{T_{i,\mathrm{gd}}(\ell_i)D}\right)_+ =  \left(\frac{T^* - 2\ell_i a_i^{-1} - (2D+1)r_{D}^{-1}}{2\ell_i a_i^{-1} D}\right)_+ , 
\end{equation}
where $(\,\cdot\,)_+=\max\{0,\cdot\}$ clips $\hat{\omega}_i$ to zero if the processing time of $i$-th user is already lower than the deadline.

{The users, which process their data before the preset deadline $T^*$, can potentially spend their residual time computing the gradients of the points offloaded by weaker users.} {The number of points that can be additionally processed by $i$-th user is given by}
\begin{equation}
	b_i = \frac{T^* - \nu_i}{T_{i,\mathrm{gd}}(1)} =\frac{a_i}{2} \left( T^* - \nu_i \right)  , \label{eq:b_i}
\end{equation}
where $\nu_i = 2(1+ D \hat{\omega}_i)\ell_i a_i^{-1}+\mathbb{I}(\hat{\omega}_i > 0)(2D+1)r_D^{-1}$ {is the processing time of $i$-th user. We further refer to the value of $b_i$ as ``gradient capacity'' of $i$-th user.}

{The minimum deadline achievable with our D2D-CFL is given by the following Lemma.}

\begin{lemma}[D2D-CFL achievable deadline]
	\label{th:s3:algorithm_finite}
	Let $W = \arg\max_i T_{i,\mathrm{gd}}(\ell_i)$ be the index of the weakest user. If the users form a connected graph, then the achievable deadline is 
	\begin{equation}
		\label{eq:optimal_deadline_finite}
		T^* = 2(1 + D \hat{\omega}_W)\ell_W a_W^{-1} + {T_d}
	\end{equation}	 
	\begin{equation}
		\label{eq:algorithm_requirement}
		\text{if}~\sum_{i=1}^N b_i \ge \sum_{i=1}^N \mathbb{I}(\hat{\omega}_i > 0),
	\end{equation}	
	where $\mathbb{I}(.)$ is the indicator function, $T_d$ is {the maximum delivery delay},  $\hat{\omega}_i $ is defined by \eqref{eq:w_i}, and $b_i$ follows from \eqref{eq:b_i}.
\end{lemma}

\begin{proof}
	Proof is given in Appendix~A.
\end{proof}

\begin{corollary}
	D2D-CFL achieves the minimum total processing time of $T^* = 0.5T_{W,\mathrm{gd}}(\ell_W)$.
\end{corollary}

\begin{proof}
	This statement holds for $r_D \to \infty,~R\to \infty$ and follows from the fact that the total processing time is bounded by the processing time of the weakest user, and $\nu_W = 0.5T_{W,\mathrm{gd}}(\ell_W)$ where $d\to\infty$ and $\hat{\omega}_W = 1$.
\end{proof}

We note that minimization of the iteration time $T^*$ does not necessarily lead to better learning convergence. In fact, decreasing $T^*$ might cause an opposite effect as coding introduces additional noise in the system. These two conflicting trends yield a tradeoff that allows one to optimize the allocations $\omega_i$ by connecting the iteration time and the learning convergence as addressed in subsection IV.C below.

\subsection{Choice of Learning Rate}
\textit{Uncoded points.} We select the learning rate of $\alpha$ based on an estimation of the Lipschitz constant $L$, where $||\nabla\!f(\boldsymbol{\beta}_1)-\nabla\!f(\boldsymbol{\beta}_2)  ||_2 \leq L || \boldsymbol{\beta}_1 - \boldsymbol{\beta}_2||_2$ for any $\boldsymbol{\beta}_1,\boldsymbol{\beta}_2 \in \mathbb{R}^d$. 
Hence, for the uncoded points, we may continue by
\begin{equation}
	\begin{array}{c}
		\!L \!=\! \underset{ \boldsymbol{\beta}_1,\boldsymbol{\beta}_2 \in \mathbb{R}^d}{ \sup } \frac{||\nabla\!f(\boldsymbol{\beta}_1)-\nabla\!f(\boldsymbol{\beta}_2)  ||_2}{|| \boldsymbol{\beta}_1 - \boldsymbol{\beta}_2||_2}\! = \!
		\underset{ \boldsymbol{\beta}_1,\boldsymbol{\beta}_2 \in \mathbb{R}^d}{ \sup }\! \frac{||\set{X}^\intercal \set{X} \left( \boldsymbol{\beta}_1 - \boldsymbol{\beta}_2 \right) ||_2} {|| \boldsymbol{\beta}_1 - \boldsymbol{\beta}_2||_2}  . \label{eq:L}
	\end{array}
\end{equation}
The above is an $\ell_2$-operator norm of the matrix $\Sigma_X = \set{X}^\intercal \set{X}$ and equals to the largest singular value of $\Sigma_X$. As singular values of a covariance matrix are its eigenvalues, the expression in \eqref{eq:L} can be estimated using various techniques for approximating spectral distributions for symmetric real-valued random matrices \cite{arnold1967asymptotic, yin1988limit, arkharov1971limit}.  Denoting the maximum eigenvalue of matrix $\set{X}^\intercal \set{X}$ as $\lambda_{\max}\left(\set{X}^\intercal \set{X}\right)$, we may continue by
\begin{equation}
	\begin{array}{c}
		{L}(d,m)  = \lambda_{\max}\left(\set{X}^\intercal \set{X} \right),
		\label{eq:L_hat}
	\end{array}
\end{equation}
where specific values of ${L}(d,m)$ can be estimated {based on the properties of streamed data.} For the case where all points are uncoded, the learning rate is selected as $\alpha(d,m) = 2\left [{{L}(d,m)} \right]^{-1}$ according to the well-known rule $\alpha = \frac{2}{L}$. Our experimental results suggest that for a wide range of parameters $d$ and $m$, the convergence provided by $\alpha(d,m)$ is either optimal or tightly close to the optimal one and can only be improved marginally.

\textit{Coded points.} If all points in the system are coded with the same coding rate $\gamma = \frac{1}{m}$, then the corresponding Lipschitz constant can be estimated as the only non-zero singular value of the matrix $(\set{G}\set{X})^\intercal \set{G}\set{X}$ since the rank of $\set{G}^\intercal\set{G}$ is constrained by $c=1$, i.e., $\text{rank}(\set{G}^\intercal\set{G}) =1$. The estimated values of the Lipschitz constant ${L}_c(d,m)$ and their functional dependency on $d$ and $m$ may also be reconstructed empirically or estimated w.r.t. the constant ${L}(d,m)$ for the uncoded case. Hence, the learning rate for the fully coded points is selected as $\alpha_c(d,m) = 2\left [{{L}_c(d,m)} \right]^{-1}$.

\textit{Partially coded points.} For the case where only a share of users decide to code their data, we choose the learning rate for each user individually as follows. Based on the share of compressed points $\hat{\omega}_i$, each user utilizes the corresponding Lipschitz constant for coded points $\alpha_c(d,m)$ and selects the learning rate ${\alpha}_{c,i} = 2\left[{L}_c(d,\hat{\ell}_i)\right]^{-1}$. The common Lipschitz constant is estimated based on the total number of uncoded data points in the system, that is, ${\alpha}_u = 2\left[{L}(d,m - \sum_{i=1}^N\hat{\ell}_i)\right]^{-1}$. Each user multiplies the partial gradient of the compressed points by the individual learning rate ${\alpha}_{c,i}$, while the partial gradients of uncoded points are weighed with the common learning rate ${\alpha}_u$. At the server, weighed compressed partial gradients are summed up, and the model is updated based on weighed compressed partial gradients and weighed uncoded partial gradients. The estimated ${\alpha}_{c,i}$ may lead to instability, especially when computed from small data batches. Therefore, we find it the best practice to divide the individual learning rate $\alpha_{c,i}$ by $2$ and by $3$ for synthetic and realistic datasets, respectively.

\begin{figure*}[!t]
	\centering
	\subfigure
	{
		\includegraphics[width=0.5\linewidth]{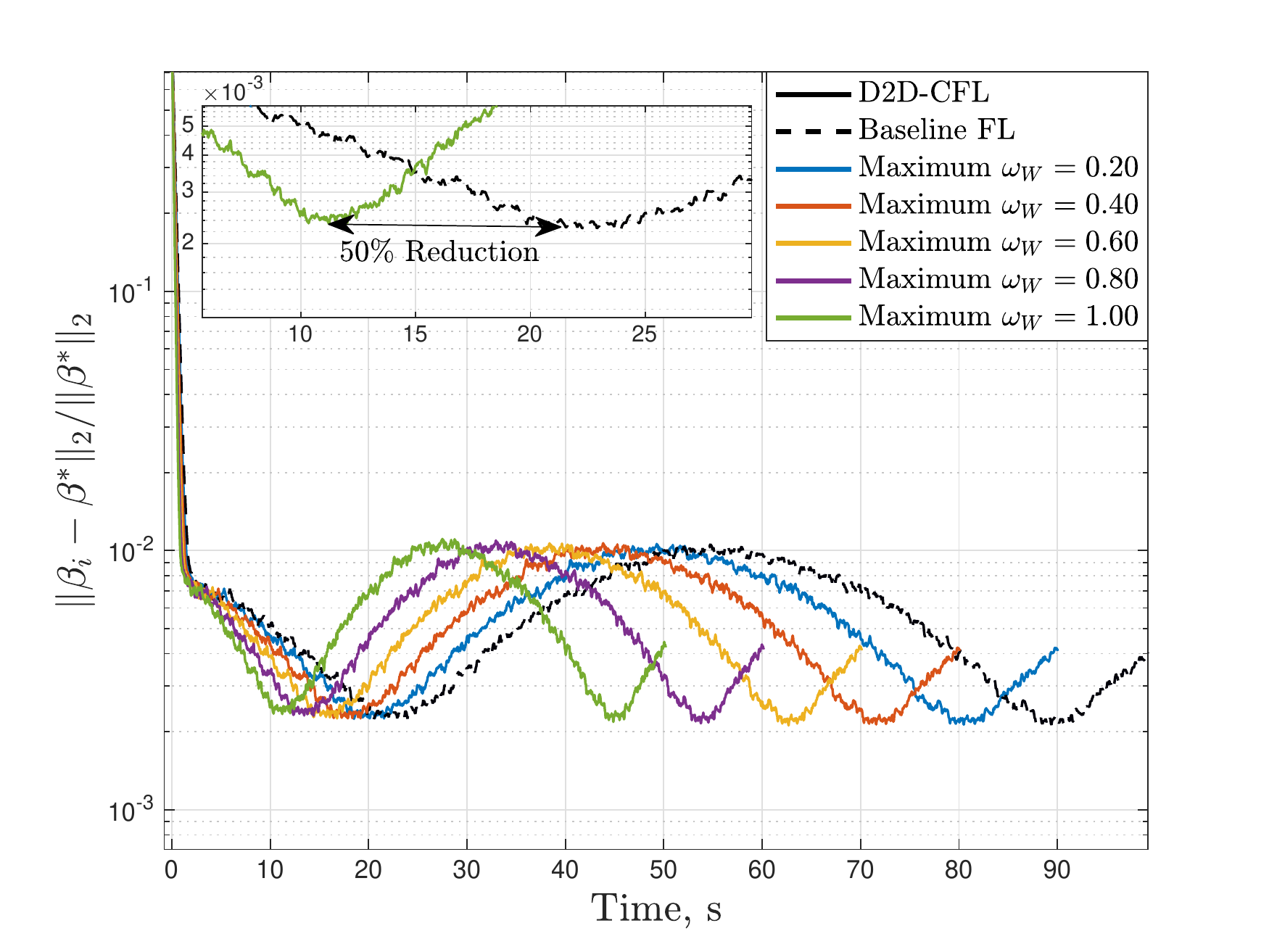}
	}~
	\subfigure
	{
		\includegraphics[width=0.5\linewidth]{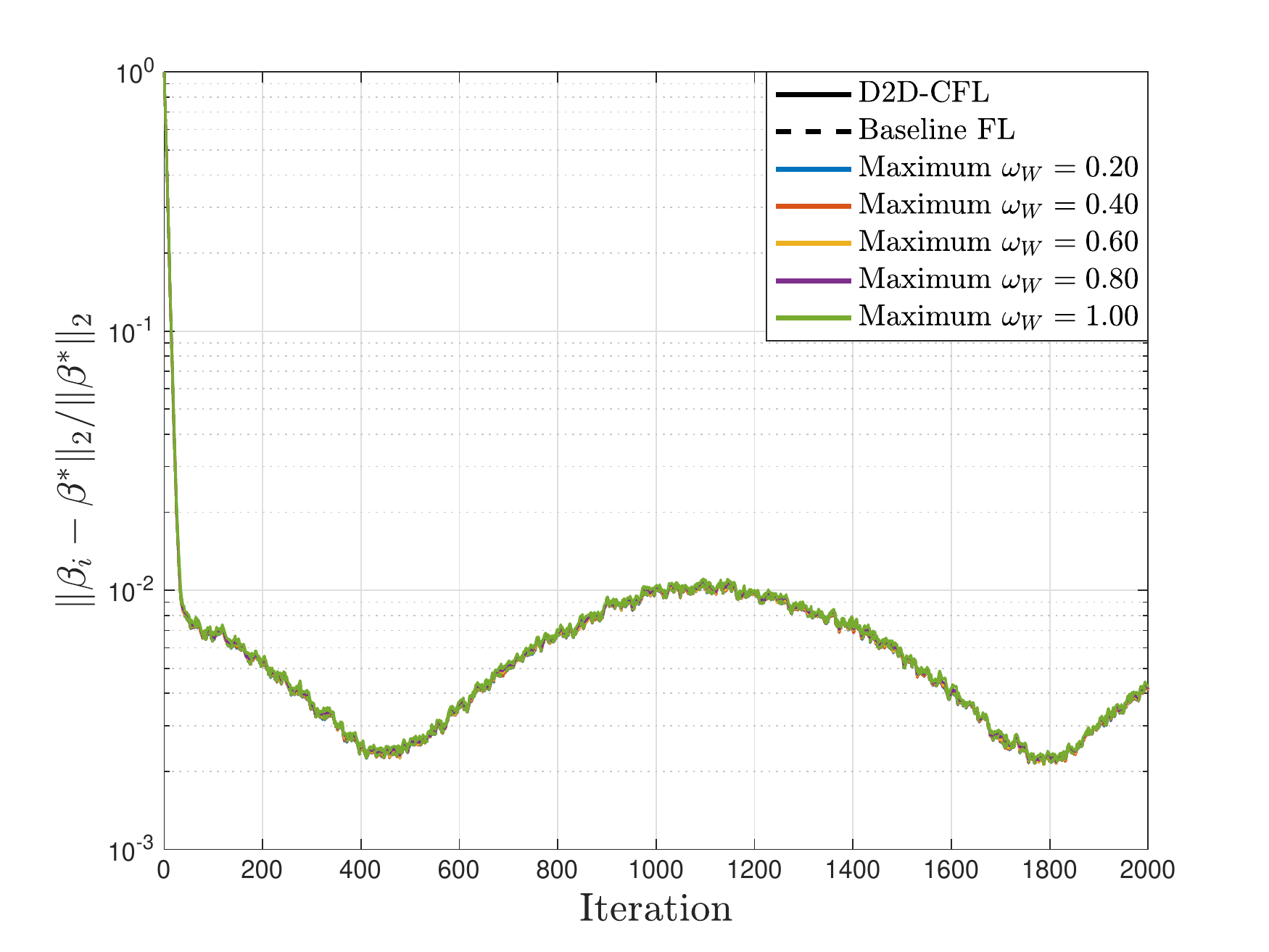}
	}
	\caption{Comparison of baseline FL and D2D-CFL in time and iterations. Dynamic model~$\boldsymbol{\beta}^*$: normalized error $\frac{||\boldsymbol{\beta}^{(t)} - \boldsymbol{\beta}^*||_2}{|| \boldsymbol{\beta}^*||_2}$ for \textit{optimal} compression rate $\gamma_i$. Improvement is up to $50\%$.}
	\vspace{-3mm}
	\label{fig:dyn_var_coding}
\end{figure*}

\begin{figure*}[ht]
	\centering
	\subfigure
	{
		\includegraphics[width=0.5\linewidth]{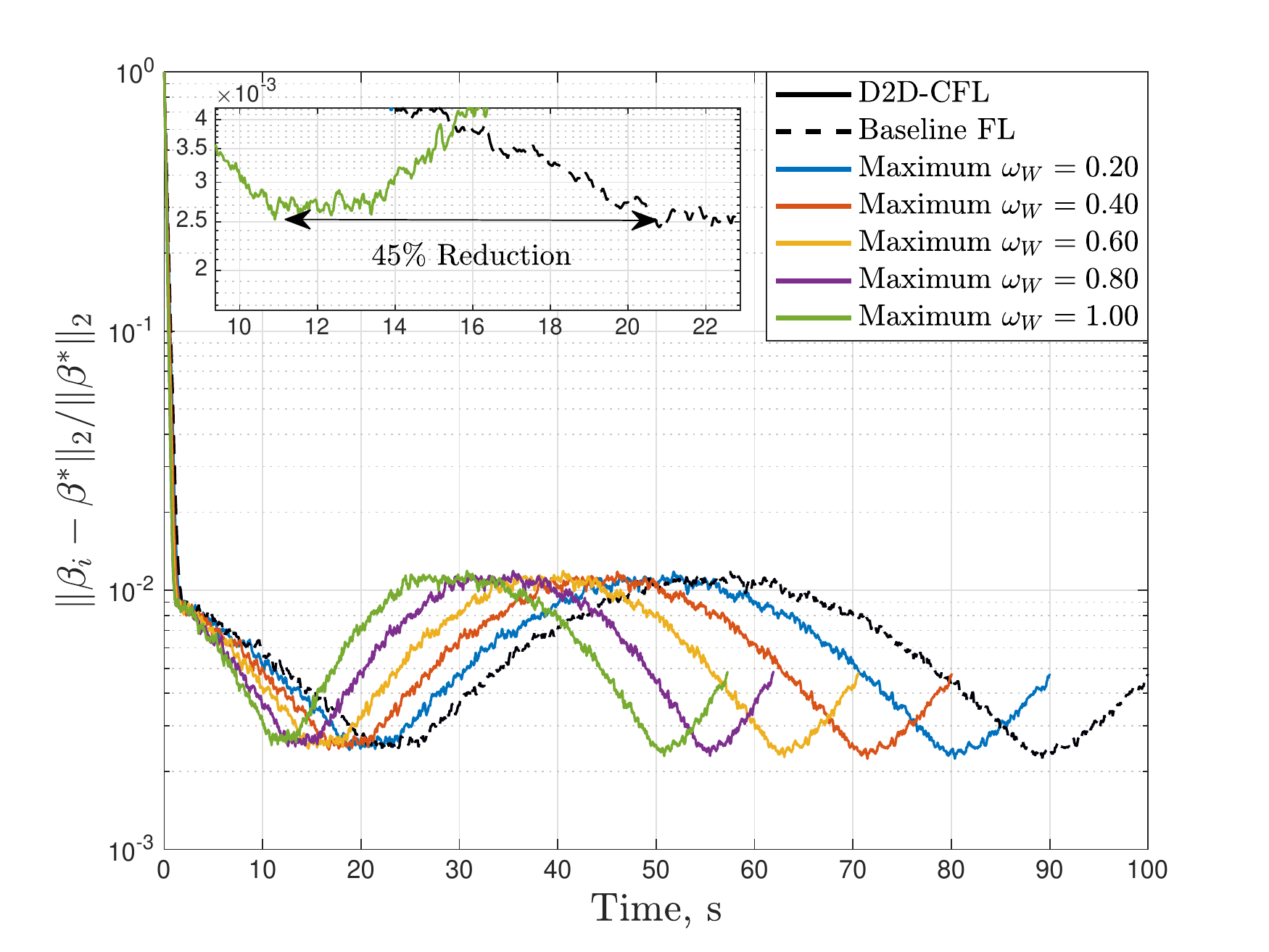}
	}~
	\subfigure
	{
		\includegraphics[width=0.5\linewidth]{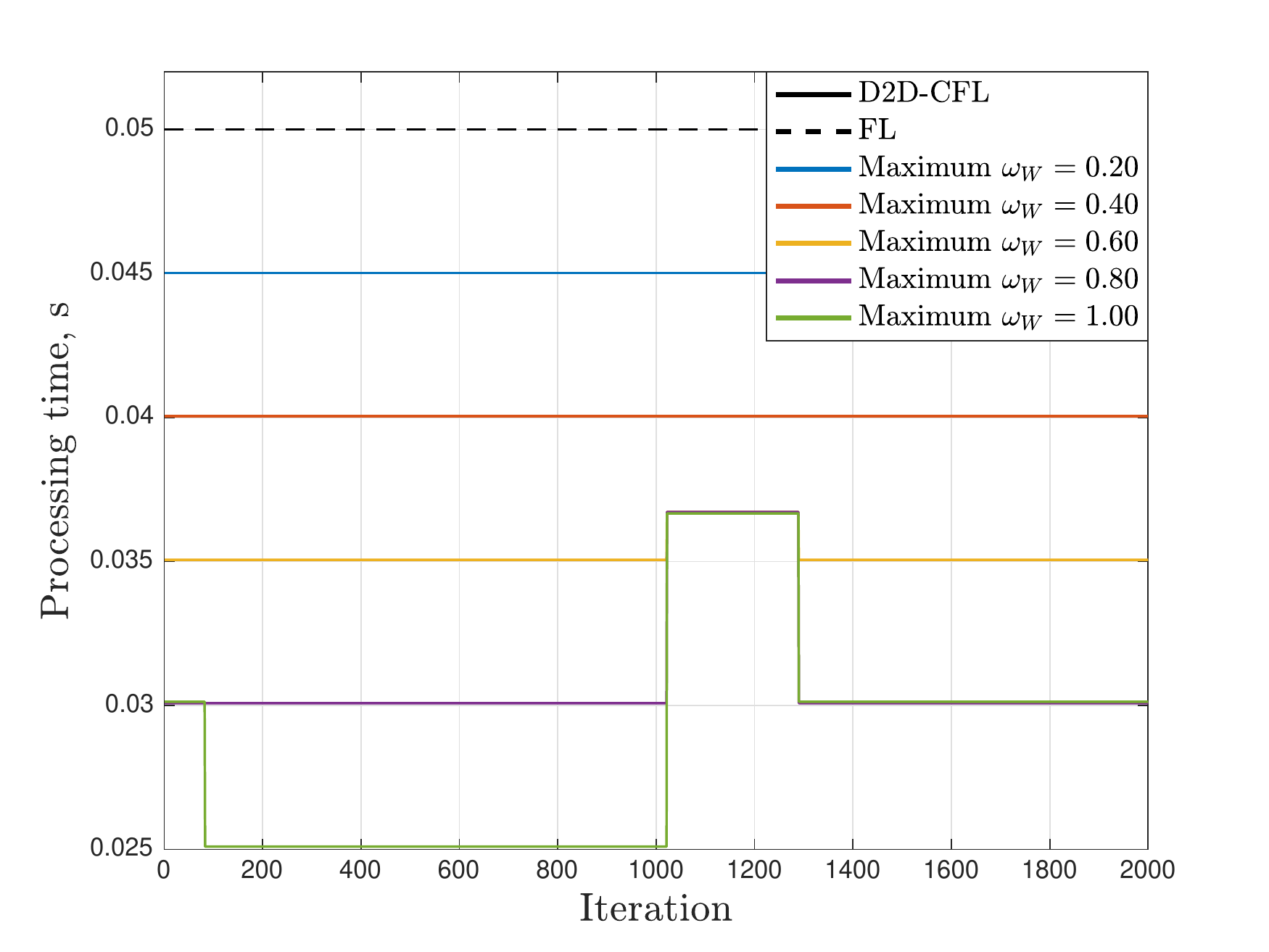}
	}
	\caption{Comparison of baseline FL and D2D-CFL in time and iterations for maximum velocity $V_{\mathrm{max}}=1$ m/s. Dynamic model~$\boldsymbol{\beta}^*$: normalized error $\frac{||\boldsymbol{\beta}^{(t)} - \boldsymbol{\beta}^*||_2}{|| \boldsymbol{\beta}^*||_2}$ and processing time for \textit{optimal} compression rate $\gamma_i$. Improvement is up to $45\%$.}
	\vspace{-3mm}
	\label{fig:mobility_low}
\end{figure*}

\begin{figure*}[!t]
	\centering
	\subfigure
	{
		\includegraphics[width=0.5\linewidth]{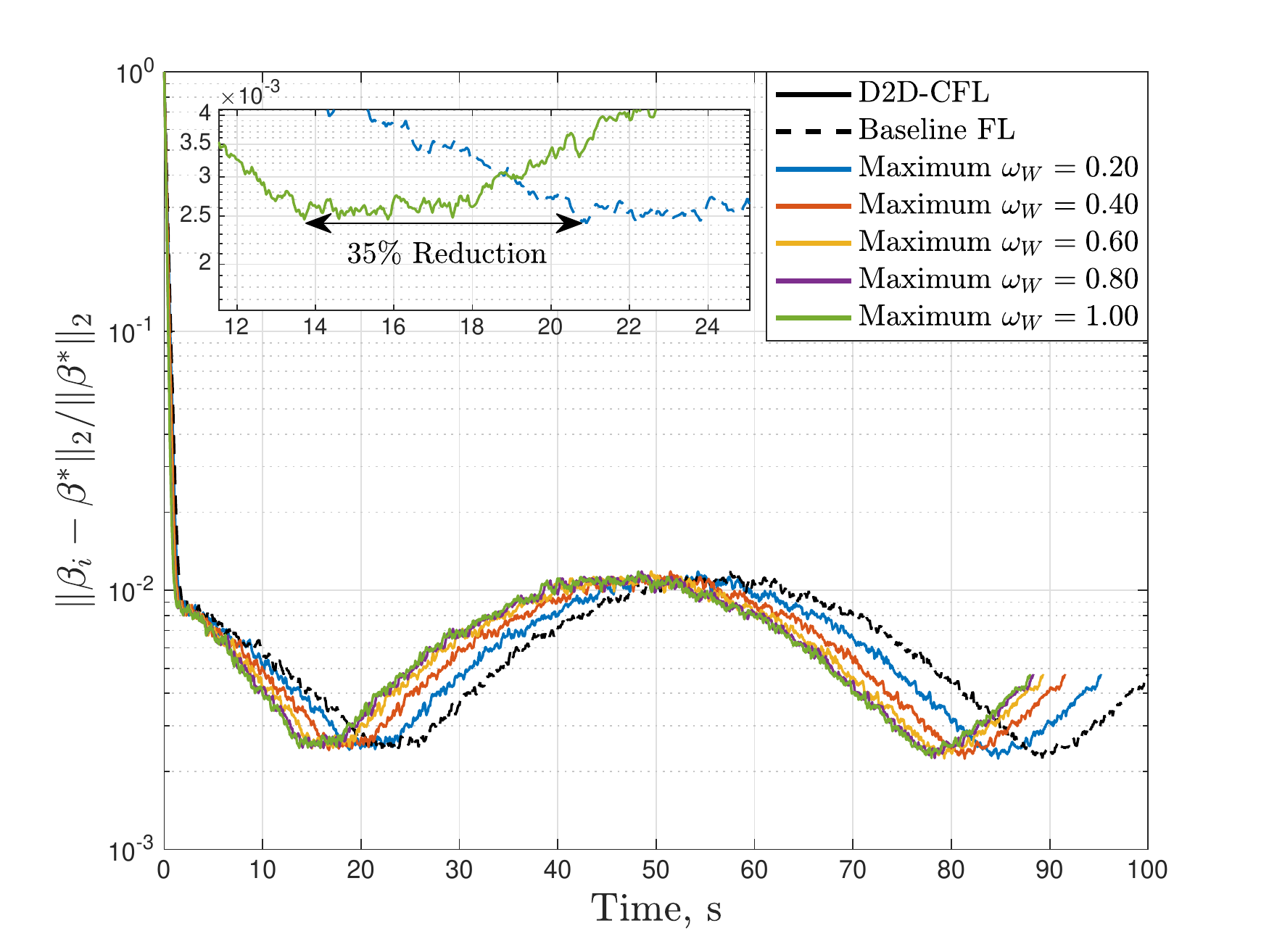}
	}~
	\subfigure
	{
		\includegraphics[width=0.5\linewidth]{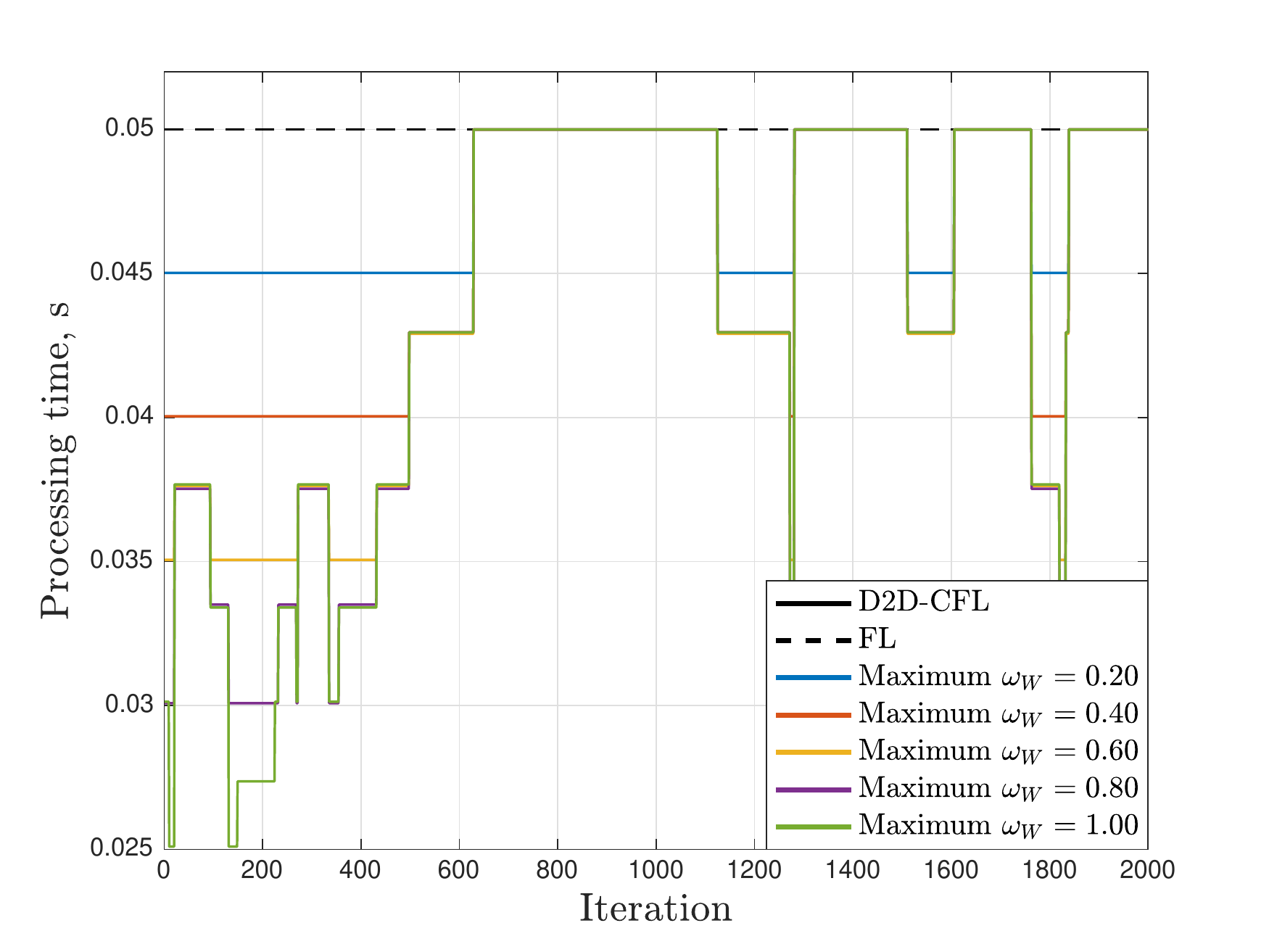}
	}
	\caption{Comparison of baseline FL and D2D-CFL in time and iterations for maximum velocity $V_{\mathrm{max}}=10$ m/s. Dynamic model~$\boldsymbol{\beta}^*$: normalized error $\frac{||\boldsymbol{\beta}^{(t)} - \boldsymbol{\beta}^*||_2}{|| \boldsymbol{\beta}^*||_2}$ and processing time for \textit{optimal} compression rate $\gamma_i$. Improvement is up to $35\%$.}
	\vspace{-3mm}
	\label{fig:mobility_high}
\end{figure*}

\subsection{Convergence Estimates for D2D-CFL}
\begin{theorem}[Approximation of error evolution without coding]
	If all points in the system are uncoded, then the evolution of the normalized regression error $\frac{||\boldsymbol{\beta}^{(t)} - \boldsymbol{\beta}^*||_2}{|| \boldsymbol{\beta}^*||_2}$ can be approximated by the following function:
	\begin{equation}
		\begin{array}{c}
			\frac{||\boldsymbol{\beta}^{(t)} - \boldsymbol{\beta}^*||_2}{|| \boldsymbol{\beta}^*||_2} = e^{-\alpha m \left(1 -\alpha \frac{d}{2} \right) t } ,
		\end{array}
	\end{equation} 
	where $\alpha$ is the learning rate, $m$ is the total number of samples, and $d$ is the number of predictors.
\end{theorem}
\begin{proof}
	Proof is given in Appendix~B.
\end{proof}

\begin{theorem}[Approximation of error evolution for fully coded option]
	If all points in the system are coded with the coding rate $\gamma$, then the evolution of the normalized regression error $\frac{||\boldsymbol{\beta}^{(t)} - \boldsymbol{\beta}^*||_2}{|| \boldsymbol{\beta}^*||_2}$ can be approximated by:
	\begin{equation}
		\begin{array}{c}
			\frac{||\boldsymbol{\beta}^{(t)} - \boldsymbol{\beta}^*||_2}{|| \boldsymbol{\beta}^*||_2} = e^{-\alpha_c m \left(1 -\alpha_c \frac{d}{2} \right) t } ,
		\end{array}
	\end{equation} 
	where $\alpha_c$ is the learning rate selected for the coded solution, $m$ is the total number of samples, and $d$ is the number of predictors.
\end{theorem}
\begin{proof}
	Proof is given in Appendix~C.
\end{proof}

\begin{corollary}
	Let $\alpha_m = \sum_{i=1}^{m-\ell_0} \alpha_{c,i} \hat{\omega}_i \ell_i + \alpha_u \ell_0$ denote the aggregate learning rate selected for the D2D-CFL solution, where $\ell_0= m- \sum \limits_{i=1}^N \hat{\omega}_i \ell_i$ is the number of uncoded points, $m$ is the total number of samples, and $d$ is the number of predictors. Then, the time evolution of the normalized regression error $\frac{||\boldsymbol{\beta}^{(t)} - \boldsymbol{\beta}^*||_2}{|| \boldsymbol{\beta}^*||_2}$ can be approximated by:
	\begin{equation}
		\begin{array}{c}
			\frac{||\boldsymbol{\beta}^{(t)} - \boldsymbol{\beta}^*||_2}{|| \boldsymbol{\beta}^*||_2} =e^{-{\alpha_m \left(1 -\alpha_m \frac{d}{2} \right)} \left[  (1 + D \hat{\omega}_W) \right] ^{-1} \frac{a_W}{2\ell_W} t } ,
		\end{array} \label{eq:condition_for_w_opt}
	\end{equation} 
	where $D = \frac{1}{2}\left(1+\frac{1}{d}\right) -1$. The shares of coded points $ \hat{\omega}_W$ and $\hat{\omega}_i$ are given by \eqref{eq:w_i}.
\end{corollary}
\begin{proof}
	Proof follows immediately from Lemma 3 and Theorem 5.
\end{proof}

The optimal share $\hat{\omega}_W$ that minimizes the total convergence time can be found via maximizing the coefficient of $t$ in \eqref{eq:condition_for_w_opt}. Hence, we may formulate the following optimization problem:
\begin{equation}
	\begin{array}{c}
		\underset{\hat{\omega}_W}{\max} \quad \alpha_m(\ell_0) \ell_0 \left(1 -\alpha_m(\ell_0) \frac{d}{2} \right) \left[  (1+D \hat{\omega}_W) \right] ^{-1} \frac{a_W}{2\ell_W}  ,
	\end{array} \label{eq:optimization_convergence}
\end{equation} 
where $\ell_0 =m- \sum \limits_{i=1}^N \hat{\omega}_i \ell_i$ is the number of uncoded samples.

For the sake of simplicity, we disregard the dependence of $\alpha_m$ on $\ell_0(\hat{\omega}_W)$ since the change in $\alpha_0(\hat{\omega}_W)$ is marginal compared to the change in the iteration time as $\hat{\omega}_W$ varies. Hence, one may rewrite the problem in \eqref{eq:optimization_convergence} as
\begin{equation}
	\begin{array}{c}
		\underset{\hat{\omega}_W}{\min} \quad \frac{ -m+ \sum \limits_{i=1}^N \left(\frac{ (1+D \hat{\omega}_W) - 2\ell_i a_i^{-1}}{2\ell_i a_i^{-1} D}\right)_+  \ell_i }{   (1+D \hat{\omega}_W) } .
	\end{array} \label{eq:optimization_convergence2}
\end{equation} 
Coupling \eqref{eq:optimization_convergence} and the feasibility condition in (\ref{eq:algorithm_requirement}), we arrive at the following Corollary.
\begin{corollary}
	The optimal share $\hat{\omega}_W$ is characterized by:
	\begin{equation*}
		\hat{\omega}_W = \arg\min\limits_{\omega_W} \frac{\sum_{i=1}^N \left(\omega_i-1\right) \ell_i}{1 + D \omega_W} + B\left(\sum_{i=1}^N \mathbb{I}(\omega_i > 0) - b_i\right)_+,
	\end{equation*}
	\noindent
	where $B \gg 1$ is the penalty and $\omega_i$ is given by (\ref{eq:w_i}).
\end{corollary}

The above optimization problem is convex, and the optimal solution can be readily found numerically. If one explicitly incorporates the dependency of $\alpha_m$ on $\ell_0$ into \eqref{eq:optimization_convergence}, the updated estimate $\hat{\omega}_i$ could lead to even better performance of the system. Our experiments indicate that varying $\hat{\omega}_i$ around this suboptimal heuristic value can only result in statistically insignificant convergence improvements.

\begin{figure*}[!h]
	\centering
	\subfigure[Comparison for equal compression rate $\gamma_i$.]
	{
		\includegraphics[width=0.5\linewidth]{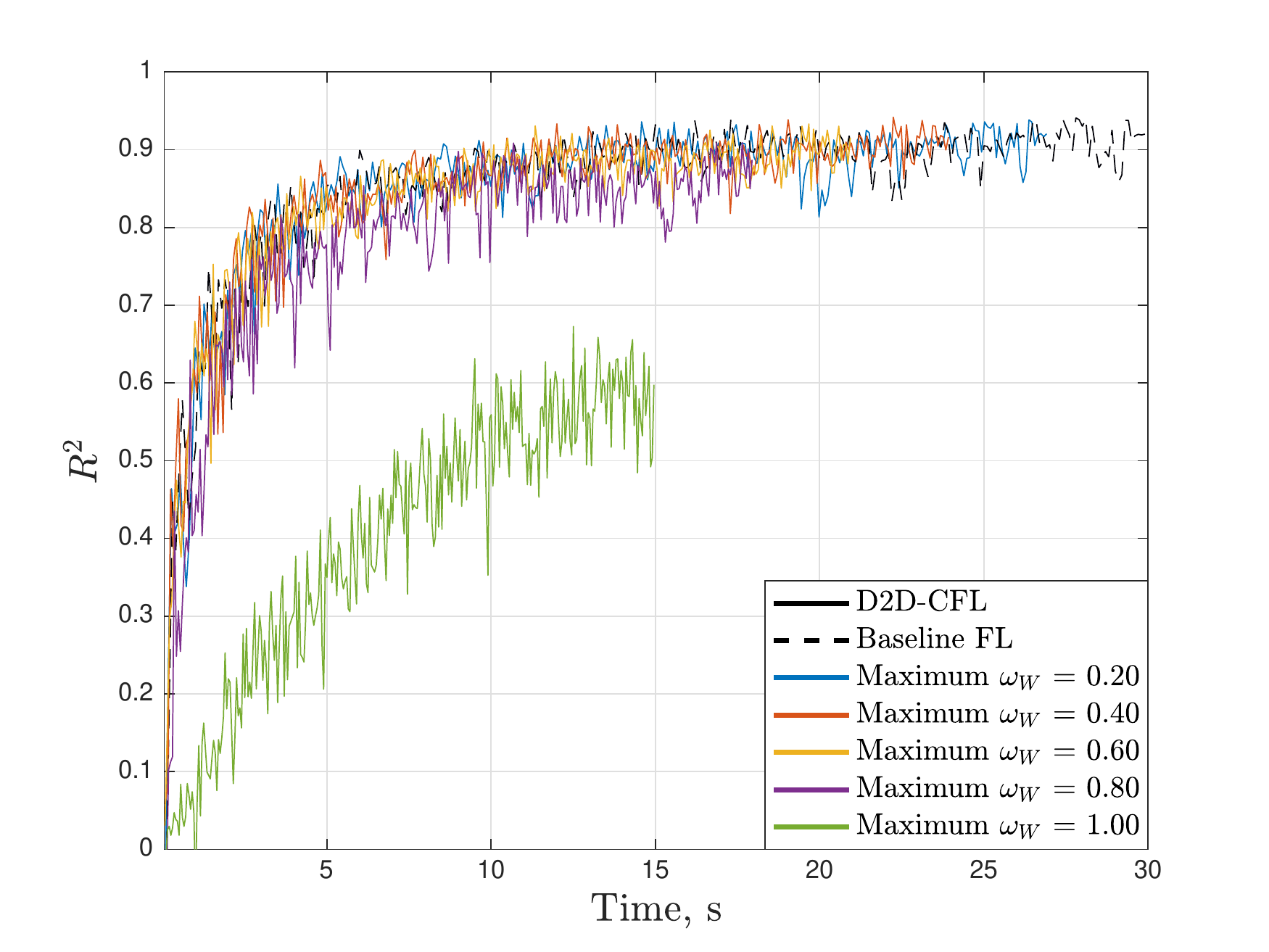}
		\label{fig:eq_coding_a}
	}~
	\subfigure[Comparison for optimal compression rate $\gamma_i$.]
	{
		\includegraphics[width=0.5\linewidth]{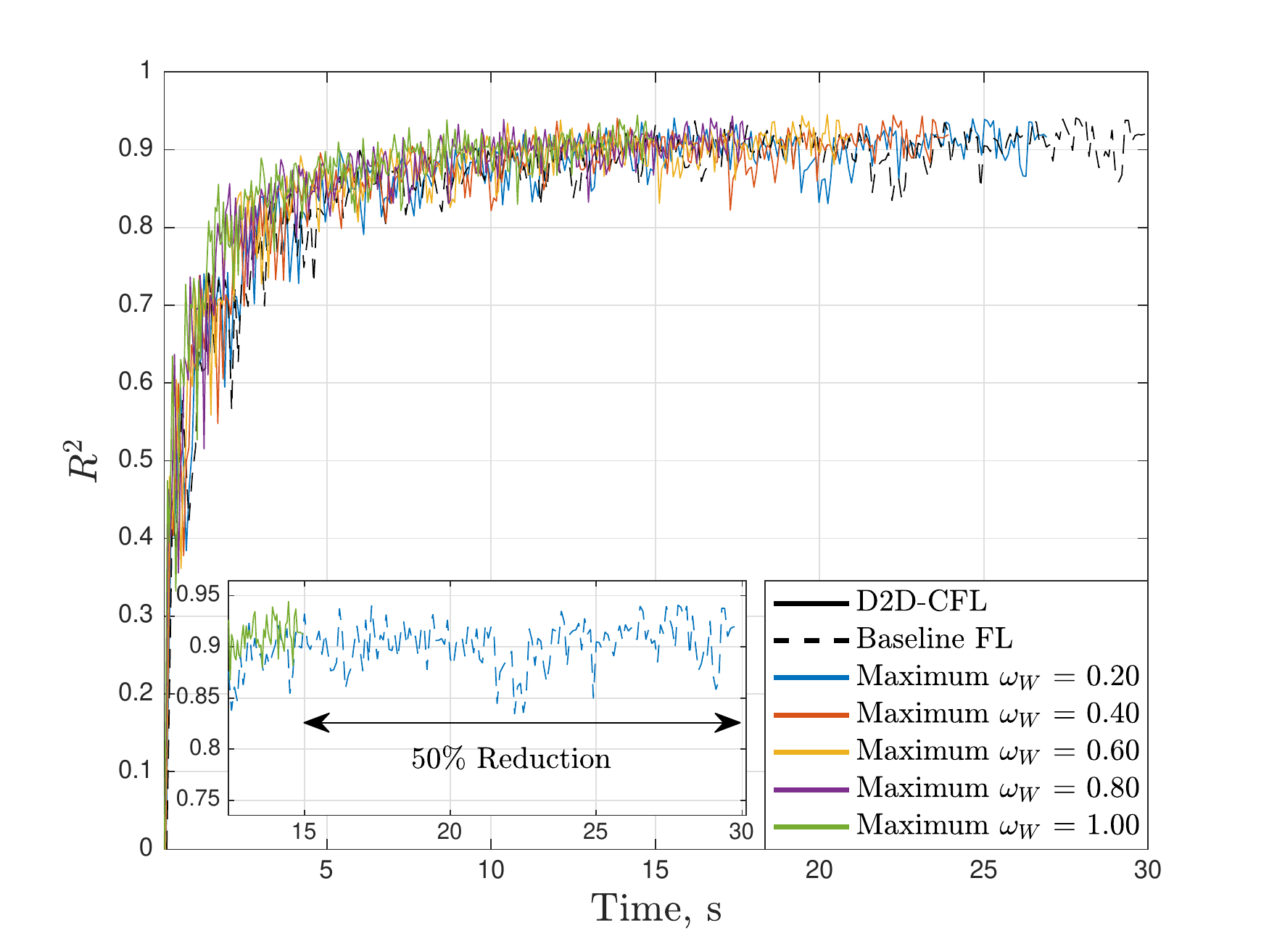}
		\label{fig:eq_coding_b}
	}
	\caption{Comparison of baseline FL and D2D-CFL in time on \textit{kin40k} dataset.}
	\vspace{-3mm}
	\label{fig:eq_coding}
\end{figure*}

\section{Selected Numerical Results}

\begin{table}[b!]\footnotesize
	\centering
	\caption{Summary of modeling parameters}
	\label{tab:gains}
	\begin{tabular}{p{1.2cm}p{6cm}|p{1.2cm}|p{1.2cm}}
		\hline
		\multirow{2}{*}{\textbf{Notation}} & \multirow{2}{*}{\textbf{Description}} & \multicolumn{2}{c}{\textbf{Value}}\\ \cline{3-4}
		&  & \textit{kin40k} & \textit{Synth.} \\
		\hline
		$N$ &Number of users &5 &25 \\
		$\ell_i$ &Batch size per user &20 &10\\
		$m$ &Number of points per iteration $N \ell_i$&100 &250 \\ \cline{3-4}
		$\rho$ &Heterogeneity factor & \multicolumn{2}{l}{0.2}\\
		$A_{\mathrm{min}}$ &Minimum computational rate &\multicolumn{2}{l}{$4\cdot 10^2$ MAC/s}\\
		$A_{\mathrm{max}}$ &Maximum computational rate, $A_{\mathrm{min}}/\rho$ &\multicolumn{2}{l}{$2\cdot 10^3$ MAC/s}\\
		$V_{\max}$ &Maximum velocity & \multicolumn{2}{l}{1 m/s, 10 m/s}\\
		$\omega_R$ &Angular rate (dynamic model) & \multicolumn{2}{l}{$4.7\cdot 10^{-2}$ rad/s}\\ \cline{3-4}
		$\sigma$ &Data noise & Not est. & $0.01$\\ \cline{3-4}
		$R_D$ &D2D proximity radius & \multicolumn{2}{l}{$4$ m}\\ \cline{3-4}
		$P_\mathrm{tx}$ &Transmit power & \multicolumn{2}{l}{10 dBm}\\
		$f$ &Carrier frequency & \multicolumn{2}{l}{60 GHz}\\
		$W$ &Bandwidth & \multicolumn{2}{l}{200 MHz}\\
		$\theta$ &Half-power beamwidth & \multicolumn{2}{l}{$\pi/9$ rad}\\
		$N_0$ &Noise power spectral density & \multicolumn{2}{l}{$-163$ dBm/Hz}\\
		\hline
	\end{tabular}
\end{table}

\subsection{Simulation Setup}
In our numerical example, the computational power distribution of the users is produced by an arithmetic sequence that increases linearly between the minimum and the maximum values $A_{\mathrm{min}}$ and $A_{\mathrm{max}}$.  
We control the computational heterogeneity by manipulating the heterogeneity factor $\rho=\frac{A_{\mathrm{min}}}{A_{\mathrm{max}}}$, meaning that low values of $\rho$ correspond to high computational heterogeneity of the system. For all experiments, we fix the minimum computational rate as  $A_{\mathrm{min}}=4\cdot 10^{2}$ MAC/s. 

{We consider {line-of-sight millimeter-wave D2D links}, assuming the free-space path loss propagation model, i.e., ${P_{\mathrm{rx}}=P_\mathrm{tx}G_\mathrm{tx}G_\mathrm{rx}\left(\!\frac{c}{4\pi f D}\!\right)^{\!2}}$,
where $P_\mathrm{rx}$ and $P_\mathrm{tx}$ are receive and transmit powers, $G_\mathrm{tx}$ and $G_\mathrm{rx}$ are directivity gains of transmit and receive antennas, $D$ is the distance, and $c$ is the speed of light. The devices utilize directional antennas with half-power beamwidth $\theta$; the corresponding directivity gains are approximated using a spherical cup area, i.e., $G_\mathrm{tx} = G_\mathrm{rx} = \frac{2}{1 - \cos{\theta}/{2}}$ {for perfectly aligned beams.}}

{The parameters of the wireless link are based on the IEEE 802.11ad standard \cite[Section 20]{ieee}. We compute the channel budget with respect to the noise factor of $10$dB and $5$dB implementation loss. A high level of the receive power on short distances allows using high-rate modulation and coding scheme (MCS) 12.3 \cite[Table 20-15]{ieee} with the $64$-QAM modulation and $5/8$ code rate. The resulting transmission rate $r_\text{D}$ is calculated as a capacity of a discrete-input continuous-output channel with an additive white Gaussian noise and bandwidth $W$.}

The user mobility is modeled according to the Random Waypoint Model as follows. {By the beginning of the training process, the two-dimensional locations of the users are uniformly distributed within the area of interest.} The movement time is divided into frames having duration of $T_F$, and each frame includes $T_F / T$ iterations, where $T$ is the duration of one iteration of the learning process. At the beginning of {each} frame, the users randomly choose their directions and velocities $V \leq V_{\max}$ and maintain the chosen parameters throughout the frame. 

\subsection{Performance Evaluation of D2D-CFL: idealistic case}
First, we illustrate the maximum achievable performance of the proposed method by considering an idealistic scenario, wherein the D2D radius tends to infinity, i.e., $R\to\infty$, and the load forwarding capacity is not limited by the graph structure constraints.

We demonstrate the advantages of D2D-CFL with the optimal coding on a synthetic dataset. We generate a linear model $\boldsymbol{\beta^*}$ and training batches $\set{X}\in\mathbb{R}^{m\times d}$ from a normal distribution. For each batch, we compute the corresponding observations $\set{y}=\set{X}\boldsymbol{\beta^*}+z$, where $z$ is a Gaussian distributed value with zero mean and standard deviation $\sigma=0.01$. 
\begin{figure}[!h]
	\centering
	\includegraphics[width=0.7\columnwidth]{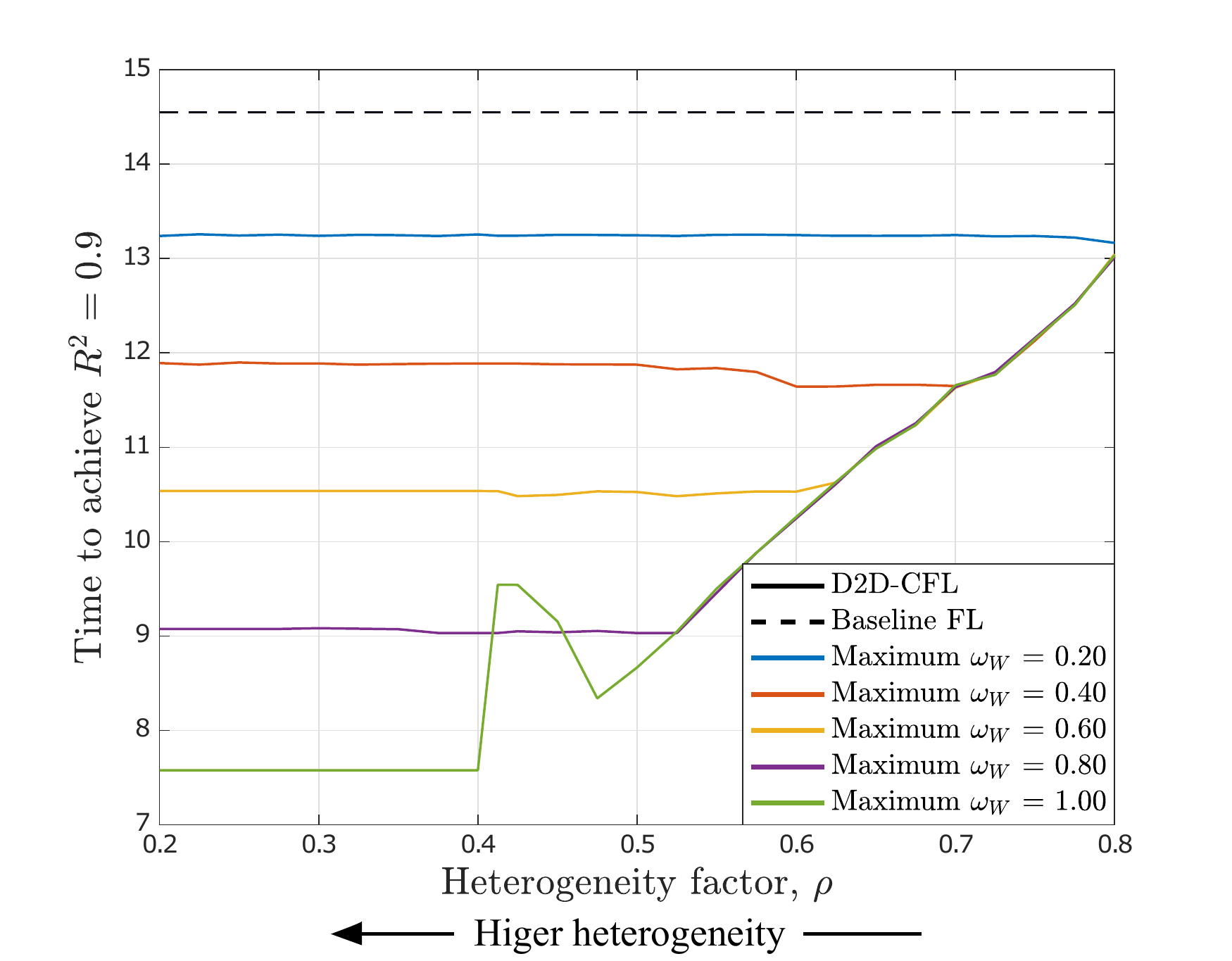}
	\caption{Comparison of baseline FL and D2D-CFL training times on \textit{kin40k} dataset for varied heterogeneity factor $\rho$.}
	\label{fig:heterogeneity}
\end{figure}
We divide $m=250$ points equally between $N=25$ users, that is, each user computes the partial gradient over $\ell_i=10$ points. To illustrate the effects of  optimal coding, we compare the performance of D2D-CFL with optimal coding rate (\ref{eq:s3a:optimal_gamma}) and fixed equal coding rates for all users. We introduce a representative level of heterogeneity by fixing the heterogeneity factor to $\rho=0.2$. Accordingly, Fig.~\ref{fig:eq_coding_synth} and \ref{fig:opt_coding_synth} demonstrate that the optimal coding reduces the training time by nearly 50\%, while the equal coding fails to demonstrate any reasonable performance gain and  provides only 8\% reduction for our experiment.

\subsection{Hyperplane Dynamics}

Further, we model the case, where the hyperplane follows a sine function with random initial phase shifts $\varphi\sim\mathrm{Uni}[0,2\pi]$ for each predictor. That is, $j$-th component of the model at $t$-th iteration has the form of $\beta^*_{j}(t) = \sin(\varphi_j + \omega_R t)$, where the angular rate $\omega_R$ is fixed for all the components. For example, Fig.~\ref{fig:dyn_var_coding} compares D2D-CFL with the optimal coding rate against the baseline for $\omega_R=4.7\cdot 10^{-2}$ rad/s. Our method shows similar performance in terms of iterations, while reducing the  training times by approximately 50\% at the error level of $2.2\cdot 10^{-3}$. Optimal coding leads to identical convergence so that the performance curves coincide.

\subsection{Impact of Direct Connectivity}

In our experiments, we construct the following mobility scenario. The movement time is divided into frames having the duration of $T_F=5$ seconds, $20$ frames in total. Each frame incorporates several iterations, while the iteration time is fixed to the total processing time of the baseline FL method. At the beginning of an iteration, the users instantly change their positions along the movement vector and perform distributed gradient computation according to the D2D-CFL solution. We assume that the transmission is instant, which in the case of a low number of users or high computational  heterogeneity may be a reasonable assumption as weaker users always transmit one point.

We employ the same dynamic model as in subsection V.D. We vary the dynamics of our model by changing the maximum velocity $V_{\max}$. The ``slow'' model with $V_{\max}=1$ m/s and the ``fast'' model with $V_{\max}=10$ m/s are presented in Fig.~\ref{fig:mobility_low} and \ref{fig:mobility_high}, respectively. Frequent fluctuations in the processing time for the ``fast'' model are due to the continuously changing connectivity graph. Despite that, the ``fast'' model always outperforms the baseline and provides $35\%$ processing time reduction at the error level of $2.5\cdot 10^{-3}$. 
 
\subsection{Realistic Dataset: kin40k}
Assuming that the users are static, in this set of experiments, we illustrate the behavior of the proposed solution on realistic datasets. We explore \textit{kin40k} dataset ($4\cdot10^4$ points, $d=8$) generated with a realistic robot arm simulator \cite[Ch. 5]{rasmussen1997evaluation}. A typical regression problem related to \textit{kin40k} dataset is to predict the distance from the robot arm end-point to a fixed position given $8$ joint angles of the arm. The problem is inherently non-linear and, therefore, one should project the original data onto a higher-dimensional feature space using kernel functions or feature mapping, which can be utilized in the case of distributed training \cite{rahimi2007random}. A popular kernel function applied to \textit{kin40k} is \textit{squared exponential} \textit{with automatic relevance determination} (SE ARD):
\begin{equation}
	\label{eq:ard_kernel}
	k(\set{x}_1,\set{x}_2)=c\cdot e^{-\frac{1}{2}(\set{x}_1-\set{x}_2)\Sigma^{-1}(\set{x}_1-\set{x}_2)^T},
\end{equation}
where $\Sigma = \mathrm{diag}\left(h_1^2,\ldots,h_d^2\right)$ is a diagonal matrix of squared length-scales $h_1^2,\ldots,h_d^2$, and $c$ is a scaling factor. The SE ARD kernel allows adjusting $h_1^2,\ldots,h_d^2$ individually for each predictor depending on its importance, or relevance to the target \cite[Sec. 1.2.3]{neal1995bayesian}. The magnitude of length-scale is inversely proportional to the relevance: varying the irrelevant features with large length scales has low impact on the prediction. As a result, the feature selection procedure is automatic and reduces to optimization of \textit{kernel hyperparameters} $\boldsymbol{\theta}=\left(h_1,\ldots,h_d,c\right)$.

\begin{figure*}[!h]
	\centering
	\subfigure[Communications efficiency of D2D-CFL for $N=5$ users.]
	{
			\includegraphics[width=0.5\linewidth]{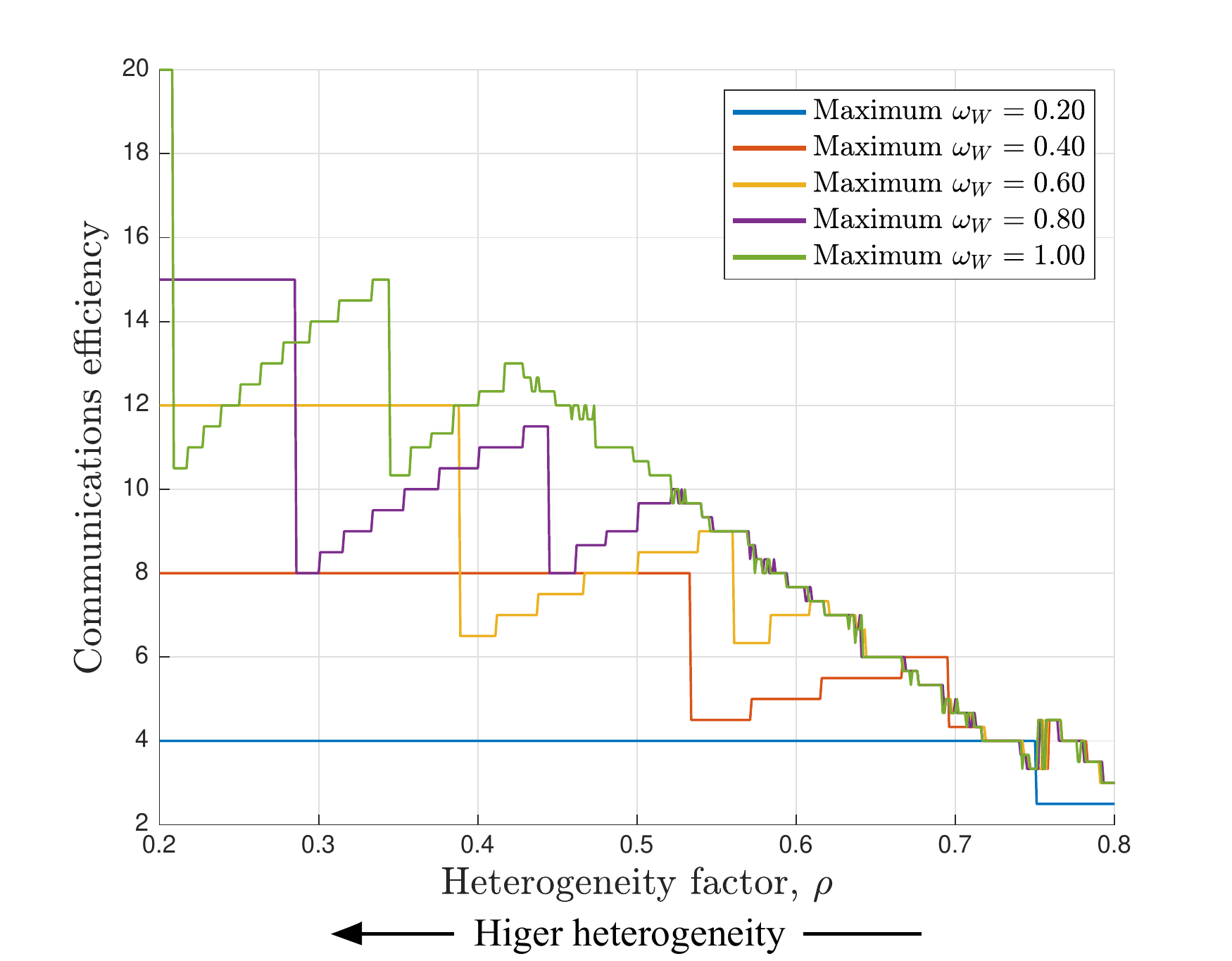}
			\label{fig:comm_eff_5}
	}~
	\subfigure[Communications efficiency of D2D-CFL for $N=50$ users.]
	{
			\includegraphics[width=0.5\linewidth]{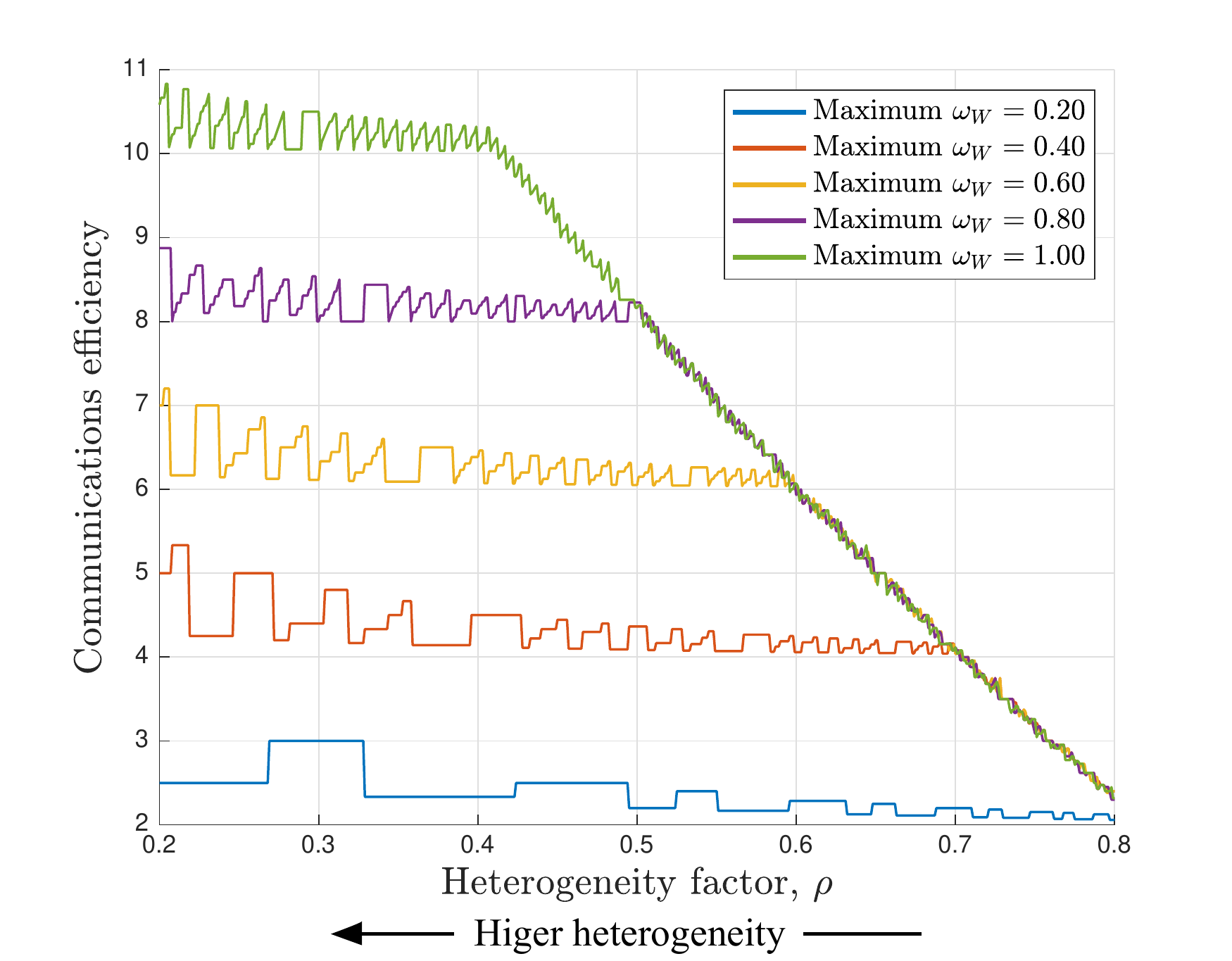}
			\label{fig:comm_eff_50}
	}
	\caption{Communications efficiency of D2D-CFL on \textit{kin40k} dataset for varied heterogeneity factor $\rho$.}
	\label{fig:comm_eff}
	\vspace{-3mm}
\end{figure*}

The positive definite kernel $k(\set{x}_1,\set{x}_2)$ and the corresponding feature mapping $\phi\colon \mathbb{R}^{1\times d}\to\mathbb{R}^{1\times \hat{d}}$ are related as $k(\set{x}_1,\set{x}_2)=\phi(\set{x}_1)\phi(\set{x}_2)^T$. Since \eqref{eq:ard_kernel} is a non-degenerate shift-invariant kernel, we approximate $\phi(\set{x})$ with a \textit{random Fourier feature mapping} (RFFM) \cite{rahimi2007random}:
\begin{equation*}
	\hat{\phi}(\set{x}) = \sqrt{\frac{2c}{\hat{d}}}\left[\cos(\set{x}\set{w}_1+\varphi_1),\ldots,\cos(\set{x}\set{w}_{\hat{d}}+\varphi_{\hat{d}})\right],
\end{equation*}
where $\set{w}_i\sim\mathcal{N}(0,\Sigma^{-1})$ is a $d$-dimensional random vector, $\varphi_i\sim\mathrm{Uni}[0,2\pi]$, and $\Sigma $ is defined in \eqref{eq:ard_kernel}. 

We assume that the first $10^4$ points of the employed dataset have been previously observed by the users and utilized to fit the hyperparameters $\boldsymbol{\theta}$. Based on \cite[Sec. 2.1.2, 5.4.1]{rasmussen2006gaussian}, we may accurately estimate $\boldsymbol{\theta}$ by maximizing the log-likelihood $\log p(\set{y}|\set{X},\boldsymbol{\theta})$ using the GPML toolbox \cite{rasmussen2010gaussian}.

The remaining $3\cdot 10^4$ points are used to train the model and compare the performance of the baseline FL with the proposed D2D-CFL. We divide the test data into $300$ batches of size $m=100$ and evenly distribute them between $N=5$ users. At each iteration, the users compute the gradient over $\ell_i=20$ points according to the proposed solution. We estimate the predictive performance of our collaboratively trained model by measuring the coefficient of determination as
\begin{center}
	\vspace{-0.8cm}
	\begin{equation*}
		R^2 = 1 - \|\set{y} - \set{X}\boldsymbol{\beta}\|_2^2 / \|\set{y} - \mathbb{E}[\set{y}]\|_2^2,
	\end{equation*}
	\vspace{-0.8cm}
\end{center}
which characterizes the goodness-of-fit using new data from the \textit{next} iteration. Predictive performance for equal and optimal choices of the compression rate $\gamma_i$ for the heterogeneity factor of $\rho=0.2$ is illustrated in Fig.~\ref{fig:eq_coding}. The bottom line in Fig.~\ref{fig:eq_coding_a} illustrates the case where the users encode their \textit{entire} training set into one point. In this situation, the noise resulting from coding cannot be compensated with clean, uncoded data, which results in substantial degradation of convergence time.

Similar to the experiment with the synthetic dataset, we also vary heterogeneity factor $\rho$ to understand the effect of system heterogeneity on the performance. In Fig.~\ref{fig:heterogeneity}, we illustrate the time required to achieve $R^2=0.90$ on \textit{kin40k} dataset; the time is averaged over the window of $10$ iterations to mitigate the presence of random effects. The curves display the maximum achievable deadline $T^*$ averaged over $10^4$ experiments for each heterogeneity factor~$\rho$. The noise caused by the high compression may degrade the performance at certain iterations, which results in a peak of the curve $\omega_W = 1$.  Such behavior is caused primarily by the quality of estimation of the learning rate for coded points $\alpha_{c,i}$ and compression of particular small data batches.

In Fig.~\ref{fig:comm_eff}, we {study} the communications efficiency of D2D-CFL by measuring the ratio of {the} total amount of data transmitted over D2D with and without the proposed compression method. The communications efficiency tends to decrease with heterogeneity factor $\rho$: the {curve} periodically drops {when} more weak users engage in offloading. By the heterogeneity factor of $\rho=0.4$, the system has exactly $\lfloor N/2 \rfloor$ weak offloading users. {W}hile their number is saturated and fixed, the total amount of offloaded data decreases, which explains {a} steady decline in the communications efficiency after $\rho=0.4$. This trend is better expressed for a large number of users as in Fig.~\ref{fig:comm_eff_50}, while for smaller numbers, the maximum communications efficiency can be achieved by varying the value of $\omega_W$ as can be concluded from Fig. \ref{fig:comm_eff_5}.

\section{Conclusion}
\vspace{-0.1cm}

In this work, we developed a novel {communication-efficient} D2D-aided {distributed learning} solution with privacy-{aware} load balancing capabilities. To accelerate the learning process, we coupled the convergence rate of distributed online regression with the optimal compression rate. We conclude with the list of key features of our method:

\begin{enumerate}
	\item The system is sensitive to the heterogeneity level, which can be directly observed by comparing results for equal and optimal compression rates in Fig.~\ref{fig:eq_coding}. The weakest user limits the total processing time, and this effect is reduced by the use of compression. {The heterogeneity level also significantly affects the communication efficiency of the proposed solution, as shown in Fig.~\ref{fig:comm_eff}.}
	\item The system performance is highly sensitive to the user connectivity level. If the users form disjoint subgraphs, the performance is limited by the weakest user \textit{and} the total gradient capacity $c_i$ of the subgraphs.
	\item Our method has the capability to capture dynamically changing models. Comparing the performance of our solution to that of the baseline FL in Fig.~\ref{fig:dyn_var_coding}, we observe that D2D-CFL not only follows the model dynamics but also introduces notable processing gains.
\end{enumerate}

In the future, we aim to extend our research by adapting D2D-CFL to, e.g., support vector machines and Bayesian models and by developing kernel selection solutions. Our study on the privacy aspects of D2D-CFL can also be extended by establishing explicit privacy guarantees or combining various differentially-private mechanisms. 

\subsection{Proof of Lemma \ref{th:s3:algorithm_finite}}
\label{th:app:algorithm_proof}

Let $a_1 = a$ and $a_2 = a + \delta$, $\delta > 0$, be the computational powers of the weakest user and a stronger arbitrary user, respectively. Without loss of generality, we assume that the users have equal numbers of points $\ell_1 = \ell_2 = \ell$ as there exists $\delta^\prime\colon 2(l_1 + \Delta)(a_2 + \delta^\prime) = 2(\ell_1)(a_2 + \delta),~\Delta > 0$. The function $\nu_i$ is linear and monotonically increasing with $\omega_i$, hence, the maximum processing time for each user is achieved for $\omega_i = 1$. Let $\omega_1 = \omega_2 = 1$, then

\begin{equation*}
	\frac{\nu_2}{\nu_1} \sim \frac{(1+D)T_{2,\mathrm{gd}}(\ell_2)}{(1+D)T_{1,\mathrm{gd}}(\ell_1)} = \frac{a}{a + \delta} < 1,
\end{equation*}

\noindent
and the total processing time of D2D-CFL is bounded by the processing time of the weakest user.

According to (\ref{eq:s3a:optimal_gamma}), it is optimal to compress any number of points into one. By definition of the gradient capacity $b_i$, $i$-th user is able to process additional $c_i$ points without increasing the total processing time. Let $K=\sum_{i=1}^N \mathbb{I}(\omega_i  > 0)$ and $J = \sum_{i=1}^N c_i$, where $\mathbb{I}(\omega_i  > 0)$ is the indicator function. Therefore, the system can process $K > J$ compressed points by distributing them across the network with the {maximum} delay of  $T_d$.

\vspace{-0.02cm}
\subsection{Proof of Theorem 5}
Let us consider the evolution of $\boldsymbol{\beta}^{(t)}$ for the case where no data is being coded in the network. The update $\boldsymbol\beta^{(t+1)} = \boldsymbol\beta^{(t)} - \alpha \nabla \! f$ can be reformulated as
\begin{equation}
\beta^{(t+1)} = \beta^{(t)} - \alpha \left ( \mathbb{E}\left [\nabla \! f^{(t)} \right] + \boldsymbol \xi^{(t)} \right), \label{eq:beta_descrete_evolution}
\end{equation}
where $ \mathbb{E}\left [\nabla \! f^{(t)} \right]$ is the expectation of the gradient and $\xi^{(t)}$ is the corresponding noise component. Here, $\set{X}$ and $\set{y}$ are the sets of the data points and observations aggregated over all users in the network. One can readily check that $\nabla \! f^{(t)} = \sum_{j=1}^{m} \nabla \! f_j^{(t)}$, where $m$ is the total number of points and $\nabla \! f_j^{(t)}$ is the gradient for an individual point $(\set{x}_j,y_j)$, $\set{x}_j\in \mathbb{R}^d$ and $y_j \in \mathbb{R}$. Therefore, the expectation of the gradient over the aggregated points translates to $ \mathbb{E}\left [\nabla \! f^{(t)} \right] =m \mathbb{E}\left [\nabla \! f_j^{(t)} \right]$. 

We continue by calculating the expectation of the gradient $\nabla \! f_j^{(t)}$ for $j$-th point as
\begin{equation}
	\mathbb{E}\left [\nabla \! f_j^{(t)} \right]  = \mathbb{E}\left [  \set{x}_j^\intercal \right( \set{x}_j \boldsymbol\beta^{(t)} - {y_j}\left) \right] 
	= \mathbb{E}\left [  \set{x}_j^\intercal \right( \set{x}_j \left( \boldsymbol\beta^{(t)} -\boldsymbol\beta^{*} \right) - {z_j}\left) \right] = \mathbb{E}\left [  \set{x}_j^\intercal  \set{x}_j \right] \left( \boldsymbol\beta^{(t)} -\boldsymbol\beta^{*} \right).
\end{equation}
Here, $z_j$ is the data noise with zero mean and standard deviation of $\sigma$. The corresponding gradient noise is given by
\begin{equation}
\begin{array}{c}
\boldsymbol \xi^{(t)}  =    \set{x}_j^\intercal  \left( \set{x}_j \boldsymbol \beta^{(t)} - y_j \right) - \mathbb{E}\left [  \set{x}_j^\intercal  \set{x}_j \right] \left( \boldsymbol\beta^{(t)} -\boldsymbol\beta^{*} \right).
 \end{array}
\end{equation}
Approximating \eqref{eq:beta_descrete_evolution} by a continuous process, we obtain
\begin{equation}
\frac{ \mathrm{d} \boldsymbol \beta^{(t)}}{ \mathrm{d}t }= - \alpha m \left ( \mathbb{E}\left [  \set{x}_j^\intercal  \set{x}_j \right] \left( \boldsymbol\beta^{(t)} -\boldsymbol\beta^{*} \right) + \boldsymbol \xi^{(t)} \right).
\end{equation}
Under the assumption on the normalization and independence of the data, we may further approximate the evolution of $\boldsymbol \beta^{(t)}$ by an ordinary differential equation (ODE) as
 \begin{equation}
\frac{ \mathrm{d} \boldsymbol \beta }{ \mathrm{d}t }= - \alpha  m\left( \boldsymbol\beta^{(t)} -\boldsymbol\beta^{*} \right) . \label{eq:beta_cont_evolution}
\end{equation}
Let the initial hyperplane value be equal to a vector of zeros, i.e., $\boldsymbol \beta^{(0)} = \set{0}$. One can readily see  that the solution to the Cauchy problem that corresponds to \eqref{eq:beta_cont_evolution} can be written as
\begin{equation}
 \boldsymbol \beta =\boldsymbol \beta^* \left( 1 - e^{-\alpha m t}\right) .
\end{equation}
Consequently, we may establish the value of the normalized error as follows:
\begin{equation}
\frac{|| \boldsymbol \beta -\boldsymbol \beta^* ||_2}{|| \boldsymbol \beta^* ||_2} = e^{-\alpha m t}
\end{equation}
where $m$ is the total number of points utilized at each iteration, and no data is being coded.

The expression in \eqref{eq:beta_descrete_evolution} may be further improved by introducing a dependency on the dimension $d$ as follows:
\begin{equation}
 \boldsymbol \beta =\boldsymbol \beta^* \left( 1 - e^{-\alpha m (1-\frac{d}{2}\alpha) t}\right) . \label{eq:beta_updated}
\end{equation}

\textbf{Remark.} We note that instead of the above simple approximation in \eqref{eq:beta_cont_evolution}, one may employ a stochastic ODE. Then, the dynamics can be described via the standard Brownian motion $\set{B}(t) \in \mathbb{R}^d$~\cite{shiryaev1999essentials} as
\begin{equation}
\mathrm{d} \boldsymbol \beta = - \alpha  m \left( \boldsymbol\beta -\boldsymbol\beta^{*} \right) \mathrm{d}t + \alpha \sigma m  \mathrm{d}\set{B}(t) . \label{eq:beta_cont_evolution_brownian}
\end{equation}
The solution of \eqref{eq:beta_cont_evolution_brownian} is a multivariate Ornstein-Ulenbeck process~\cite{uhlenbeck1930theory} and, thus, $\boldsymbol \beta$ can also be approximated by
\begin{equation}
 \boldsymbol \beta =\boldsymbol \beta^* \left( 1 - e^{-\alpha m t}\right) + \alpha\sigma m \int \limits ^t_0 e^{\alpha m (\tau -t)} \mathrm{d}\set{B}(\tau),
\end{equation}
where the integral is a stochastic It\^{o} integral with respect to the Brownian motion~\cite{shiryaev1999essentials}.

\subsection{Proof of Theorem 6}
Let us consider the case where the data of all users are coded. We rely on the results of Lemma \ref{lemma:optimal_gamma} and let $c = 1$ for all the users. In particular, a tagged user utilizes a random matrix $\set{G} \in \mathbb{R}^{1 \times n} = (g_{1},..., g_n)$ and transforms $\ell$ data points into a coded point $\set{\tilde x}= (\sum_{j=1}^{\ell} g_j x_{j,1}, ..., \sum_{j=1}^{\ell} g_j x_{j,d})$ and a coded observation ${\tilde y}= \sum_{j=1}^{\ell} g_j y_{j}$. The corresponding data noise can be expressed as $\tilde z =  \sum_{j=1}^{\ell} g_j z_{j}$, and, hence, $\mathbb{E} [\tilde z] = 0$ and $\text{var}[\tilde z] = l\sigma^2$. 

Here, we follow the proof of Theorem 6. The regression model is updated according to \eqref{eq:beta_descrete_evolution} with the only difference that all the gradients are calculated based on the coded data. Therefore, the gradient at the server is a sum of $N$ gradients of the individual coded points, i.e., $\nabla \! \tilde f^{(t)} =\sum_{i=1}^{N} \nabla \! \tilde f_i^{(t)}$. The corresponding expectation of the gradient for $i$-th coded point is given by
\begin{equation}
		\mathbb{E}\left [\nabla \! \tilde f_i^{(t)} \right]  = \mathbb{E}\left [  \set{\tilde x}_i^\intercal \right( \set{\tilde x}_i \boldsymbol\beta^{(t)} - {\tilde y_i}\left) \right] =
		\mathbb{E}\left [  \set{\tilde x}_i^\intercal  \set{\tilde x}_i \right( \boldsymbol\beta^{(t)}-\boldsymbol\beta^{*} \left) -  \set{\tilde x}_i {\tilde z_i} \right] = 
		\mathbb{E}\left [ \set{\tilde x}_i^\intercal  \set{\tilde x}_i  \right] \left( \boldsymbol\beta^{(t)}-\boldsymbol\beta^{*} \right),
\end{equation}		
where $z_i$ is the noise of the coded data. The diagonal elements of $ \set{\tilde x}_i^\intercal  \set{\tilde x}_i $ incorporate the sum $\sum_{j=1}^{\ell} g_j^2 x_{jk}^2$, where $k$ is the number of a column. Given the properties of $g_j$ and $x_{jk}$, we may conclude that the expectation of the diagonal elements equals to~$\ell$. The expectation of the non-diagonal elements is zero. Consequently, $\mathbb{E}\left [\nabla \! \tilde f_i^{(t)} \right]  = l\left( \boldsymbol\beta^{(t)}-\boldsymbol\beta^{*} \right)$ and the full gradient may be rewritten as $\nabla \! \tilde f^{(t)} = m \left( \boldsymbol\beta^{(t)}-\boldsymbol\beta^{*} \right)$, where $m =\sum_{i=1}^{N} \ell_i $ is the total number of uncoded points.

One may, thus, conclude that the regression model evolves similarly to that in the uncoded case, i.e., as
\begin{equation}
 \boldsymbol \beta =\boldsymbol \beta^* \left( 1 - e^{-\alpha_c m t}\right) ,
\end{equation}
where $\alpha_c$ is the corresponding learning rate for the coded data. Hence, the value of the normalized error may be derived as
\begin{equation}
\frac{|| \boldsymbol \beta -\boldsymbol \beta^* ||_2}{|| \boldsymbol \beta^* ||_2} = e^{-\alpha_c m t} .
\end{equation}
An improved estimate of \eqref{eq:beta_updated} may be rewritten as
\begin{equation}
 \boldsymbol \beta =\boldsymbol \beta^* \left( 1 - e^{-\alpha_c m (1-\frac{d}{2}\alpha_c) t}\right) . \label{eq:beta_updated_coded}
\end{equation}

\normalem
\bibliographystyle{plain}
\bibliography{references.bib}
	
\end{document}